\documentclass[10pt,journal,compsoc]{IEEEtran}
\ifCLASSOPTIONcompsoc
  \usepackage[nocompress]{cite}
\else
  \usepackage{cite}
\fi
\ifCLASSINFOpdf
\else
\fi

\usepackage[noend]{algpseudocode}
\usepackage{algorithm}
\usepackage{algorithmicx}
\usepackage{multirow}
\usepackage{microtype}
\usepackage{graphicx}
\usepackage[normalem]{ulem}
\usepackage{makecell}
\useunder{\uline}{\ul}{}
\usepackage{booktabs}
\usepackage{subfigure}
\usepackage{amsmath}
\usepackage{amssymb}
\usepackage{color,xcolor}
\usepackage{url}

\newcommand{\modelName}{GAPA}

\begin{document}
%
\title{Automatic Context Pattern Generation \\ for Entity Set Expansion}
%
%
%
%

\author{Yinghui Li,
        Shulin Huang,
        Xinwei Zhang,
        Qingyu Zhou,
        Yangning Li, 
        Ruiyang Liu, \\
        Yunbo Cao,
        Hai-Tao Zheng ~\IEEEmembership{Member,~IEEE},
        and Ying Shen ~\IEEEmembership{Senior Member,~IEEE}
\IEEEcompsocitemizethanks{
\IEEEcompsocthanksitem Yinghui Li, Shulin Huang, Yangning Li, and Hai-Tao Zheng are with Shenzhen International Graduate School, Tsinghua University, Shenzhen, Guangdong 518055, China, and also with Peng Cheng Laboratory, Shenzhen, Guangdong 518066, China. E-mail: \{liyinghu20, sl-huang21, liyn20\} @mails.tsinghua.edu.cn, zheng.haitao@sz.tsinghua.edu.cn.\protect\\
\IEEEcompsocthanksitem Xinwei Zhang and Ying Shen are with the School of Intelligent Systems Engineering, Sun Yat-Sen University, Guangzhou, Guangdong 510275, China. E-mail: zhangxw57@mail2.sysu.edu.cn, sheny76@mail.sysu.edu.cn.\protect \\
\IEEEcompsocthanksitem Qingyu Zhou and Yunbo Cao are with the Tencent Cloud Xiaowei, Beijing 100190, China. E-mail: qyzhgm@gmail.com, yunbocao@tencent.com.\protect \\
\IEEEcompsocthanksitem Ruiyang Liu is with the Department of Computer Science and Technology, Tsinghua University, Beijing 100190, China. E-mail: lry20@mails.tsinghua.edu.cn.\protect \\
}
\thanks{Manuscript received 17 July 2022; revised 8 Mar 2023. \\
Yinghui Li and Shulin Huang contribute equally for this work. \\
Corresponding authors: Hai-Tao Zheng and Ying Shen.
}
}

%
%

\markboth{IEEE TRANSACTIONS ON KNOWLEDGE AND DATA ENGINEERING}%
{Li \MakeLowercase{\textit{et al.}}: Automatic Context Pattern Generation for Entity Set Expansion}
%



\IEEEtitleabstractindextext{%
\begin{abstract}
Entity Set Expansion (ESE) is a valuable task that aims to find entities of the target semantic class described by given seed entities. 
Various Natural Language Processing (NLP) and Information Retrieval (IR) downstream applications have benefited from ESE due to its ability to discover knowledge. 
Although existing corpus-based ESE methods have achieved great progress, they still rely on corpora with high-quality entity information annotated, because most of them need to obtain the context patterns through the position of the entity in a sentence.
Therefore, the quality of the given corpora and their entity annotation has become the bottleneck that limits the performance of such methods.
To overcome this dilemma and make the ESE models free from the dependence on entity annotation, our work aims to explore a new ESE paradigm, namely corpus-independent ESE.
Specifically, we devise a context pattern generation module that utilizes autoregressive language models (e.g., GPT-2) to automatically generate high-quality context patterns for entities. 
In addition, we propose the \modelName{}, a novel ESE framework that leverages the aforementioned \textbf{G}ener\textbf{A}ted \textbf{PA}tterns to expand target entities. 
Extensive experiments and detailed analyses on three widely used datasets demonstrate the effectiveness of our method. 
All the codes of our experiments are available at \url{https://github.com/geekjuruo/GAPA}.
\end{abstract}

}

\maketitle

\IEEEdisplaynontitleabstractindextext

%
\IEEEpeerreviewmaketitle

\IEEEraisesectionheading{\section{Introduction}
\label{Sec:Introduction}}

\IEEEPARstart{E}{ntity Set Expansion} (ESE) - i.e., aiming to expand complete entities that belong to the same semantic class as a few seed entities - is a promising task in IR and NLP~\cite{shi2014probabilistic, zhang2017entity, zhao2018entity}. For example, given the input seed entity set \{\emph{“China”}, \emph{“Japan”}, \emph{“Canada”}\}, an ESE model is expected to output more new entities (e.g., “\emph{Germany}”, “\emph{Australia}”, “\emph{Singapore}”, ...) that all belong to the same \texttt{Country} class as the seed entities. Thanks to its ability to automatically discover knowledge and mine semantics, the ESE task can benefit kinds of IR and NLP downstream applications, such as Web Search~\cite{chen2016long}, Knowledge Graph Completion~\cite{shi2021entity}, and Question Answering~\cite{wang-etal-2008-automatic}.

\begin{figure*}[h]
\centering
\includegraphics[width=0.75\textwidth]{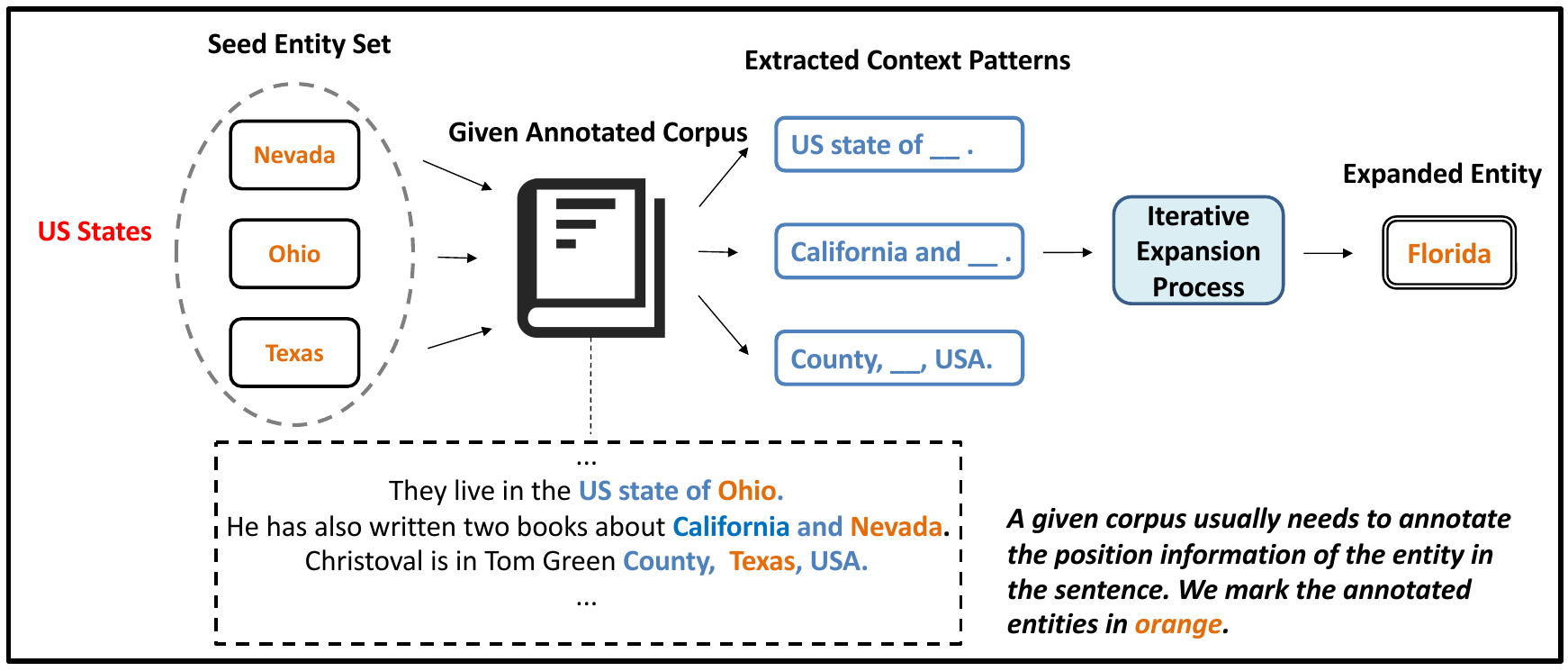}
\caption{An example showing the expansion process of the traditional corpus-based ESE methods.}
\label{Intro_Figure}
\end{figure*}

In recent years, various corpus-based bootstrapping methods have been studied and became mainstream solutions for ESE~\cite{zhao2017entity, yan2019learning, yan2020end, yan-etal-2021-progressive}. 
Given a specific corpus w
ith entities annotated, these methods mainly bootstrap the seed entity set by iteratively selecting context patterns and ranking expanded entities~\cite{velardi-etal-2013-ontolearn, rong2016egoset, mamou-etal-2018-term,huang2020guiding}. 
Taking Figure~\ref{Intro_Figure} as an example, when expanding the seed entity set \{\emph{“Nevada”}, \emph{“Ohio”}, \emph{“Texas”}\} which belongs to the \texttt{US States} class, corpus-based ESE methods will extract related context patterns from the given annotated corpus, such as “County, \underline{\ \ }, USA.”. Then iterative expansion is performed according to similarity between the context and entity representations.
However, the biggest constraint faced by these corpus-based methods is that their performance relies heavily on the entity annotation quality of the given corpus. 
When there is no given corpus or the entity annotation of the given corpus is very noisy (which is very common and normal in real-world application scenarios), the entity expansion performance of these corpus-based ESE methods will undoubtedly be greatly reduced, because it will be difficult for them to extract high-quality context patterns.
Therefore, how to make the ESE methods get rid of the dependence on the entity annotation of the corpus is a problem worthy of studying in ESE.
To address this problem, our work brings the ESE task into a new paradigm, namely corpus-independent ESE which does not need to be given a specific corpus with entity annotations.

In this study, we propose to empower ESE with context patterns automatically generated from the autoregressive language models, such as GPT-2. 
Intuitively, given an entity, the language generation model can utilize this entity as guiding text to automatically generate the corresponding context (i.e., prev-text and next-text of entities).
The context automatically generated by pre-trained language model can naturally be regarded as a context pattern that the ESE model can make use of.
Note that most of the widely used language generation models are well pre-trained without any artificially designed features or entity annotation, so that the context patterns they generate are without the help of human involvement and corpora annotation.
Furthermore, after obtaining automatically generated context patterns, we can utilize them to obtain entity's context representations for measuring entity similarity, based on the important and reasonable assumption that “similar entities should have similar context”~\cite{zhang2018entity,barrena-etal-2016-alleviating}.

Motivated by the above intuition, we propose a novel iterative ESE framework that consists of three modules: 
(1) The first, \emph{supervision signal enhancement} module, updates initial seed/candidate entity sets according to the entity similarity which is calculated from the entity representations. 
(2) The second, \emph{context pattern generation} module, utilizes two separate GPT-2 models in opposite directions to automatically generate amounts of context patterns of entities (e.g., “Bill Gates is [MASK]'s founder.” is one context pattern of the entity “Microsoft”).
(3) The third, \emph{generated patterns guided expansion} process, scores each entity in the candidate set according to its context similarity with seed entities and adds top-ranked entities into the seed set iteratively.
As our proposed method automatically generates context patterns for input entities, our framework does not require any entity annotation information and ESE task-related corpora, so we declare it is corpus-independent for the ESE task.

In summary, the major contributions of this paper are: 
\begin{enumerate}
    \item We first apply reverse GPT-2 to generate the prev-text of entities and automatically derive high-quality context patterns for the ESE task, thus freeing our proposed model from the reliance on the corpora with annotated entities.
    \item We propose a novel ESE framework, \modelName{}, which can utilize the automatically \textbf{G}ener\textbf{A}ted \textbf{PA}tterns to iteratively expand target entities effectively.
    \item Extensive experiments and detailed analyses show that \modelName{} outperforms previous state-of-the-art methods without any manually designed patterns.
\end{enumerate}

The rest of the article is organized as follows. Section~\ref{Sec:Related_Work} provides a comprehensive review of related work on entity set expansion and autoregressive language models. Section~\ref{Sec:Methodology} describes the technology details of our proposed \modelName{}. Experimental settings and analyses are presented in Section~\ref{Sec:Experiments}. Finally, we conclude our work and give an outlook on its future directions in Section~\ref{Sec:Conclusion}.

\section{Related Work}
\label{Sec:Related_Work}
\subsection{Entity Set Expansion}
Entity Set Expansion (ESE) task has attracted extensive research efforts due to its practical importance and broad application prospects~\cite{pantel2009web, shen2020taxoexpan, yu2020learning}. 
Benefitting from its ability to mine semantics and discover knowledge, the ESE task can be used for many downstream tasks or applications, such as taxonomy construction~\cite{velardi-etal-2013-ontolearn}, Query Optimization~\cite{pantel2009web} and News Recommendation~\cite{zhao2017entity}. 
Early works in ESE mainly focus on web-based methods, such as SEAL~\cite{wang2007language} and Google Sets~\cite{tong2008system}, which require search engines and web pages as external online resources for expansion. 
Due to the low efficiency of web-based methods, more recent studies have paid more attention to corpus-based methods~\cite{kushilevitz-etal-2020-two,yu2019corpus, zhang-etal-2020-empower}. Among various corpus-based methods, the iterative bootstrapping methods are the most mainstream solutions~\cite{zupon-etal-2019-lightly, yan2020end, yan-etal-2021-progressive}. 
These bootstrapping methods aim to bootstrap the seed entity by iteratively selecting context patterns and ranking expanded entities. 
To accurately extract context patterns, these methods require that the given corpus must contain annotations of where entities occur in sentences. 
For example, the previous state-of-the-art ESE methods CGExpan~\cite{zhang-etal-2020-empower} and ProbExpan~\cite{li2022contrastive} both need annotated corpus in their methods. 
CGExpan needs to have the entity position of a sentence, so as to achieve the contextual representation of the entity by BERT, and ProbExpan needs to mask the entity in the sentence to pre-train the entity-level masked language model.
In addition, there is a long-term problem of \emph{sparse supervision} in the ESE task, because it tends to have very few seed entities as input~\cite{yan2019learning}.
SynSetExpan~\cite{shen-etal-2020-synsetexpan} proposes to leverage the task of synonym discovery to give ESE model more supervision signal, thereby improving the performance of the ESE model. 
Different from SynSetExpan, we propose to directly use the entity representations to measure the similarity between entities, and then update the initial seed/candidate sets, so as to achieve the purpose of supervision signal enhancement.

\subsection{Autoregressive Language Models} 
Recently, various pre-trained language models have been applied in various natural language processing tasks and gained good performance~\cite{qiu2020pre, dong2021survey, min2021recent}. 
The mainstream pre-trained language models can be divided into two categories, namely autoencoder language models and autoregressive language models. Common autoencoder language models include BERT~\cite{devlin-etal-2019-bert}, T5~\cite{raffel2019exploring}, RoBERTa~\cite{liu2019roberta}, UniLM~\cite{dong2019unified} and XLM~\cite{lample2019cross}. The typical representative of them, BERT, cannot directly deal with text generation tasks due to its masking pre-training strategy and Transformer-Encoder structure~\cite{vaswani2017attention}.
Therefore, most autoregressive language models are naturally suitable for various natural language generation tasks, such as GPT-1~\cite{radford2018improving}, GPT-2~\cite{radford2019language} and ULMFiT~\cite{howard2018universal}, because of their unbiased estimation of joint probability and consideration of the correlation between predicted tokens.
But in this autoregressive paradigm, language models can only generate the next-text based on the prev-text, that is, generate text in the direction from left to right. The reason for this limitation is that the joint probability derived by the autoregressive language models is decomposed according to the normal text sequence order.
To the best of our knowledge, our work is the first to apply autoregressive language models to generate prev-text from right to left, and verify its feasibility and effectiveness empirically.

\begin{figure*}[]
\centering
\includegraphics[width=0.75\textwidth]{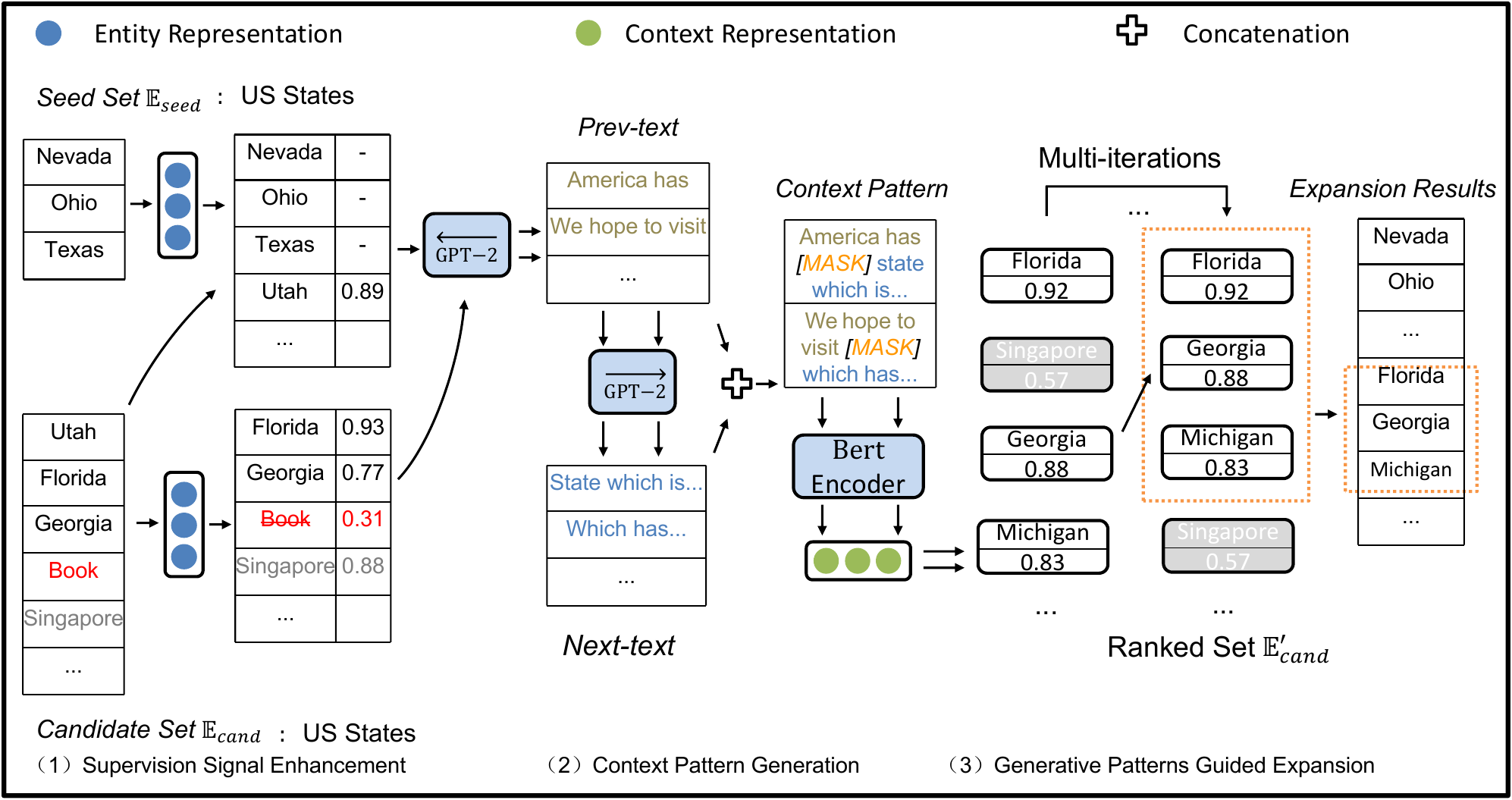}
\caption{Overview of our proposed \modelName{} framework. We update the initial seed/candidate sets according to the similarity of entity representations. Automatically generate the prev-text and next-text of entities respectively through two opposite GPT-2 models, and then concatenate them to get the context patterns. According to the similarity of context representations obtained by BERT, we iteratively select proper entities from the candidate set to add into the seed set, thus resulting in ideal target expansion results.}
\label{Method_Figure}
\end{figure*}

\section{Methodology}
 \label{Sec:Methodology}
In this section, to describe the details of our method more clearly, we first formulate the ESE task and define some key concepts of ESE (Section~\ref{Sec:Problem_Formulation}).
Specially, we will firstly introduce the mechanism of the supervision signal enhancement module (Section~\ref{Sec:Seed_Entities_Enhancement}) that we propose for the sparse supervision problem in ESE.
Then we will illustrate how we use autoregressive language models to automatically generate the context patterns of entities (Section~\ref{Sec:Context_Patterns_Generation}).
Finally, we will describe the generated patterns guided expansion process (Section~\ref{Sec:Generated_Patterns_Guided_Expansion}).
The overview of our proposed method is shown in Figure~\ref{Method_Figure} and we will summarize our proposed method in Section~\ref{Sec:Summary_Of_Method}.
Besides, we will provide a detailed example explanation for the Figure~\ref{Method_Figure} in Section~\ref{Sec:Summary_Of_Method}.

\subsection{Problem Formulation}
\label{Sec:Problem_Formulation}
\noindent\textbf{Entity.} An entity is something that is distinguishable and exists independently in real world, such as \emph{“China”} and \emph{“United States”}.

\noindent\textbf{Context Pattern.} An entity's context pattern contains a \textbf{prev-text} and a \textbf{next-text}. For example, for the sentence \emph{“We all know that Beijing is the capital of China.”} which contains the entity \emph{“Beijing”}, \emph{“We all know that *”} is the prev-text and \emph{“* is the capital of China.”} is the next-text of \emph{“Beijing”}.

\noindent\textbf{Semantic Class.} A semantic class can classify entities semantically. For instance, \emph{“Japan”} belongs to the \texttt{Country} class and \emph{“Microsoft”} belongs to the \texttt{Company} class. 

\noindent\textbf{Seed Entity Set.} A small set of few entities belonging to the same semantic class. In our work, the size of the initial seed entity set is typically 3.

\noindent\textbf{Candidate Entity Set.} For a specific ESE dataset, the candidate entity set is a large set containing all entities of all semantic classes in this dataset.

\noindent\textbf{ESE Task.} The inputs are a complete set of candidate entities and a small set of seed entities (e.g., \{\emph{“United States”}, \emph{“Japan”}, \emph{“Canada”}\}) belonging to the same semantic class (i.e., \texttt{Country}), the ideal output is a ranked list of other \texttt{Country} entities from the candidate set, such as \emph{“Korea”}, \emph{“India”}, and \emph{“Singapore”}.

\subsection{Supervision Signal Enhancement}
\label{Sec:Seed_Entities_Enhancement}
Because the number of initial seed entities is very small (usually only 3-5 entities are initialized as the seed set), while the number of candidate entities is often hundreds or thousands, this naturally poses the \emph{sparse supervision} challenge. Intuitively, if there are more seed entities and fewer candidate entities, the supervision signal will be strengthened. Based on this intuition, we propose the supervision signal enhancement module to automatically increase seed entities and reduce candidate entities. 

Given a small initial seed entity set $\mathbb{E}_{\text{seed}}$ and a large candidate entity set $\mathbb{E}_{\text{cand}}$, whether an entity should be added to the seed set or removed from the candidate set is judged according to its similarity to the initial seed entity set. Specifically, the entity similarity $\text{score}_{e}$ between a candidate entity $e_c \in \mathbb{E}_{\text{cand}}$ and $\mathbb{E}_{\text{seed}}$ is measured by computing the cosine distance between entity representations as follow:
\begin{equation}
    \text{score}_{e}(e_c) = \frac{1}{|\mathbb{E}_{\text{seed}}|}\sum_{e_s \in \mathbb{E}_{\text{seed}}} \cos{(\textbf{r}_{e}(e_c),\textbf{r}_{e}(e_s))},
    \label{equ_entity_score}
\end{equation}
where $|\mathbb{E}_{\text{seed}}|$ is the size of $\mathbb{E}_{\text{seed}}$ and $\cos{(x,y)}=\frac{x\cdot y}{||x||_2||y||_2}$ is cosine distance between two entity representations. The operator $\textbf{r}_{e}(e)$ represents the entity representation, which is defined as:
\begin{equation}
    \textbf{r}_{e}(e) = \text{Glove}(e),
\end{equation}
where $\text{Glove}$ means the pre-trained word vectors. For an entity composed of multiple words, if their concatenation can be directly found in Glove's vocabulary, their concatenated word's vector is used. Otherwise, we average the word vectors of each word as the representation of this entity.

So far, we can judge the similarity between the candidate entity and the seed entity set according to the similarity score (i.e., Equation~\ref{equ_entity_score}). 
If the candidate entity has a high similarity score with the seed entity set, it should be added to the seed set to increase the number of seed entities. It is worth noting that under the setting of ESE, it is necessary to keep the seed set and the candidate set without overlapping entities, so we need to remove these entities with high similarity from the candidate set at the same time. Conversely, if the candidate entity has a low similarity score to the seed entity set, it should be removed from the candidate entity set. Practically, we set upper and lower thresholds, i.e., $\text{thr}_u$ and $\text{thr}_l$, to distinguish different entities. Lager seed entity set $\mathbb{E}_{\text{seed}}^{'}$ and smaller candidate entity set $\mathbb{E}_{\text{cand}}^{'}$ are achieved as:
\begin{equation}
\label{update_seed}
 \mathbb{E}_{\text{seed}}^{'} = \mathbb{E}_{\text{seed}} \cup \{e | e \in \mathbb{E}_{\text{cand}}, \ \ \text{score}_{e}(e) \geq  \text{thr}_u \}, 
\end{equation}
\begin{equation}
\label{update_cand}
    \mathbb{E}_{\text{cand}}^{'} = \{ e | e \in \mathbb{E}_{\text{cand}}, \ \  \text{thr}_u \textgreater \text{score}_{e}(e) \geq \text{thr}_l \}. 
\end{equation}

By adding more similar entities in the seed set and reducing dissimilar entities from the candidate set, we finally effectively alleviate the effect of sparse supervision in ESE. In addition, a stronger set of seed entities is also beneficial to improve the quality of the generated context patterns later.

\subsection{Context Pattern Generation}
\label{Sec:Context_Patterns_Generation}
The context pattern generation module inputs seed/candidate sets and generates context patterns for entities. 
We construct this module by utilizing two separate GPT-2 models in opposite directions to generate the prev-text and next-text of entities.

As mentioned in Section~\ref{Sec:Related_Work}, traditional autoregressive language models, such as GPT-2, can only generate the next-text from left to right. Therefore, to efficiently obtain the prev-next of entities, we pre-train a reverse GPT-2 model in a simple training way. Specifically, we pre-process the regular corpus to reverse all the sentences and then use these reverse corpora to pre-train the GPT-2 model.
Apart from that, the training strategy and loss function design for training reverse GPT-2 are not different from the training process of regular GPT-2. In other words, we pre-train GPT-2 with the reverse corpus to give it the ability to generate prev-text from right to left. For the convenience, we denote the reverse GPT-2 as $\overleftarrow{\text{GPT-2}}$ and the regular GPT-2 as $\overrightarrow{\text{GPT-2}}$ in our paper.

First, for an entity $e$, we use $e$ as the guiding text and $\overleftarrow{\text{GPT-2}}$ to generate the reverse text $\overleftarrow{\text{prev}}$:
\begin{equation}
    \overleftarrow{\text{prev}} = \overleftarrow{\text{GPT-2}}(e).
\end{equation}
Note that the word order of $\overleftarrow{\text{prev}}$ is reversed at this time, we can very simply reverse it to $\overrightarrow{\text{prev}}$. Then we use $\overrightarrow{\text{prev}}$ and $e$ as the guiding text for $\overrightarrow{\text{GPT-2}}$ to generate the next-text $\overrightarrow{\text{next}}$:
\begin{equation}
    \overrightarrow{\text{next}} = \overrightarrow{\text{GPT-2}}([\overrightarrow{\text{prev}} ; e ]),
\end{equation}
where $[\cdot ; \cdot]$ denotes concatenation. Finally, combining the prev-text and next-text of $e$, we get one context pattern of $e$ :
\begin{equation}
    \textbf{c}(e) = [\overrightarrow{\text{prev}};[\text{MASK}];\overrightarrow{\text{next}}],
    \label{Equ:Pattern}
\end{equation}
where $[\text{MASK}]$ is the masked token of BERT~\cite{devlin-etal-2019-bert} which will be used to get the context representation of entities through BERT. In particular, in our framework, a [MASK] represents an entire entity regardless of how many words the entity consists of. The reason we select BERT as our encoder is that BERT is widely used as the encoder in previous advanced ESE work and other NLP tasks.

An important problem worth considering in the above procedure is the choice of GPT-2 decoding strategy. In text generation tasks, GPT-2 often uses decoding strategies such as Greedy Search~\cite{germann2003greedy}, Beam Search~\cite{vijayakumar2016diverse}, and Top-K Sampling~\cite{fan2018hierarchical}. The typical characteristic of Greedy Search and Beam Search is that when the pre-training parameters of GPT-2 and guiding text are fixed, the generated text of each time is not very different. Therefore, in order to ensure the diversity of the context patterns we generate for every entity, we choose Top-K Sampling as the decoding strategy. 

\subsection{Generated Patterns Guided Expansion}
\label{Sec:Generated_Patterns_Guided_Expansion}
For every entity in $\mathbb{E}_{\text{seed}}^{'}$ and $\mathbb{E}_{\text{cand}}^{'}$, we automatically generate $m$ context patterns through the context pattern generation module. Supported by the theoretical assumption that “similar entities should have similar contexts”, whether a candidate entity should be expanded is judged according to its context similarity to the seed set. Specifically, we measure the context similarity $\text{score}_{c}$ between a candidate entity $e_c \in \mathbb{E}_{\text{cand}}^{'}$ and $\mathbb{E}_{\text{seed}}^{'}$  by computing the cosine distance between context representations as follow:
\begin{equation}
    \text{score}_{c}(e_c) = \frac{1}{|\mathbb{E}_{\text{seed}}^{'}|}\sum_{e_s \in \mathbb{E}_{\text{seed}}^{'}} \cos{(\textbf{r}_{c}(e_c),\textbf{r}_{c}(e_s))},
\end{equation}
where $\textbf{r}_{c}(e)$ represents the average of the entity's $m$ context representations derived from BERT, which is defined as:
\begin{equation}
    \textbf{r}_{c}(e) = \frac{1}{m}\sum_{i=0}^{m-1} \text{BERT}(\textbf{c}_{i}(e)),
\end{equation}
where $\textbf{c}_{i}(e)$ means i-th context pattern of entity $e$, which is defined in Equation~\ref{Equ:Pattern}.

After obtaining the context similarity score between each entity in $\mathbb{E}_{\text{cand}}^{'}$ and $\mathbb{E}_{\text{seed}}^{'}$, our model selects the candidate entities with the top-3 scores and add them to $\mathbb{E}_{\text{seed}}^{'}$ at each iteration. Such iterations are continued until the number of entities in $\mathbb{E}_{\text{seed}}^{'}$ reaches the target expansion number. Through this iterative generated patterns guided expansion, we will get the final ideal target entities.

\subsection{Summary of Methodology}
\label{Sec:Summary_Of_Method}
Starting with a small seed entity set and a large candidate entity set, we first use the Glove as the entity representations, and update the initial seed/candidate sets according to the similarity of entity representations, which alleviates the sparse supervision problem in ESE. Secondly, we automatically generate context patterns for entities using two opposite GPT-2 models, thus making our method less dependent on annotated corpora. Then, we utilize BERT to encode the generated context patterns to obtain context representations. Finally, we iteratively rank the candidate set according to the context similarity score and select top-ranked entities into the seed set, thus achieving the purpose of the ESE task.

To clarify our proposed method more concretely, we now give an example which is illustrated in Figure~\ref{Method_Figure}. Given the seed set \{\emph{“Nevada”}, \emph{“Ohio”}, \emph{“Texas”}\} and a large candidate set, we expect our model to expand more \texttt{US States} entities into the initial set. 
Firstly, \modelName{} calculates the similarity between candidate entities and seed entities according to the entity representations. Based on the entity similarity score, \modelName{} updates the initial set and candidate set, such as adding \emph{“Utah”} with higher score to the seed set and removing \emph{“Microsoft”} with lower score from the candidate set. 
Secondly, we use $\overleftarrow{\text{GPT-2}}$ and $\overrightarrow{\text{GPT-2}}$ to generate various prev-text and next-text for entities. And we concatenate the prev-text and next-text to form the context patterns, such as \emph{“America has [MASK] state which is ...”}. 
Thirdly, we utilize BERT to encode the context patterns and get the context representations of entities. Furthermore, \modelName{} measures the similarity of entities by computing the similarity of context, so as to iteratively add more similar entities to the seed set. Taking an iteration shown in the Figure~\ref{Method_Figure} as an example, we select \emph{“Florida”}, \emph{“Georgia”}, \emph{“Michigan”} from the candidate set and add them into the seed set. This iteration continues until the size of the seed set expands to the target number.

\begin{table*}[h]
    \centering
    \caption{MAP@K(10/20/50) of different methods. All baseline results are directly from other published papers. We underline the previous state-of-the-art performance on three datasets for convenient comparison.}
    \begin{tabular}{lccccccccc}
    \toprule \multirow{2}{*} { \textbf{Methods} } & \multicolumn{3}{c} { \textbf{Wiki} } & \multicolumn{3}{c} { \textbf{APR} } & \multicolumn{3}{c} { \textbf{SE2} } \\ \cmidrule(r){2-4} \cmidrule(r){5-7}\cmidrule(r){8-10}
    & MAP@10 & MAP@20 & MAP@50 & MAP@ 10 & MAP@20 & MAP@50 & MAP@10 & MAP@20 & MAP@50 \\
    \midrule Egoset~\cite{rong2016egoset} & 0.904 & 0.877 & 0.745 & 0.758 & 0.710 & 0.570 & 0.583 & 0.533 & 0.433 \\
    SetExpan~\cite{shen2017setexpan} & 0.944 & 0.921 & 0.720 & 0.789 & 0.763 & 0.639 & 0.473 & 0.418 & 0.341 \\
    SetExpander~\cite{mamou-etal-2018-term} & 0.499 & 0.439 & 0.321 & 0.287 & 0.208 & 0.120 & 0.520 & 0.475 & 0.397 \\
    CaSE~\cite{yu2019corpus} & 0.897 & 0.806 & 0.588 & 0.619 & 0.494 & 0.330 & 0.534 & 0.497 & 0.420 \\
    Set-CoExpan~\cite{huang2020guiding} & 0.976 & 0.964 & 0.905 & 0.933 & 0.915 & 0.830 & - & - & -  \\
    CGExpan~\cite{zhang-etal-2020-empower}  & \underline{0.995} & 0.978 & 0.902 & 0.992 & \underline{0.990} & 0.955 & 0.601 & 0.543 & 0.438 \\
    SynSetExpan~\cite{shen-etal-2020-synsetexpan} & 0.991 & 0.978 & 0.904 & 0.985 & \underline{0.990} & \underline{0.960} & 0.628 & 0.584 & 0.502 \\
    ProbExpan~\cite{li2022contrastive} & \underline{0.995} & \underline{0.982} & \underline{0.926} & \underline{0.993} & \underline{0.990} & 0.934 & \underline{0.683} & \underline{0.633} & \underline{0.541} \\
    \midrule
    \modelName{}~(BERT-base) & \textbf{1.000} & \textbf{1.000} & 0.971 & \textbf{1.000} & \textbf{1.000} & 0.985 & \textbf{0.688} & 0.640 & 0.540 \\
    \modelName{}~(BERT-wwm) & \textbf{1.000} & \textbf{1.000} & \textbf{0.974} & \textbf{1.000} & \textbf{1.000} & \textbf{0.990} & \textbf{0.688} & \textbf{0.641} & \textbf{0.542} \\
    
    \midrule
    \end{tabular}
    
    \label{tab:allresult}
\end{table*}

\section{Experiments}
\label{Sec:Experiments}

\subsection{Datasets}
To ensure fairness, in addition to two widely used datasets (i.e., Wiki/APR) in previous works~\cite{shen2017setexpan,zhang-etal-2020-empower,shen-etal-2020-synsetexpan}, we also choose another larger and more challenging SE2 dataset~\cite{shen-etal-2020-synsetexpan}.
Following previous works which used Wiki/APR datasets, in our experiments, each semantic class has 5 seed sets and each seed set has 3 entity queries.
The details of our used datasets are as follows:
\begin{enumerate}
    \item \textbf{Wiki}~\cite{zhang-etal-2020-empower}, which is from a subset of English Wikipedia. It contains 8 semantic classes, including \texttt{China Provinces}, \texttt{Companies}, \texttt{Countries}, \texttt{Diseases}, \texttt{Parties}, \texttt{Sports Leagues}, \texttt{TV Channels} and \texttt{US States}. 
    \item \textbf{APR}~\cite{zhang-etal-2020-empower}, which consists of all the 2015 year's news articles published by Associated Press and Reuters. It contains 3 semantic classes, including \texttt{Countries}, \texttt{Parties} and \texttt{US States}. 
    \item \textbf{SE2}~\cite{shen-etal-2020-synsetexpan}, the largest ESE benchmark to date, which has 60 semantic classes and 1200 seed entity queries.
\end{enumerate}
Additionally, we use the Wikipedia 20171201 dump\footnote{The raw corpus we use for pre-training the reverse GPT-2 can be directly downloaded from https://archive.org/details/enwiki-20171201.} as the pre-training corpus (we denote this part of data as Reverse-Wikipedia in our paper) for the reverse GPT-2 model (i.e., $\overleftarrow{\text{GPT-2}}$) in the context pattern generation module.
It is worth noting that the Reverse-Wikipedia is the only training data we use, which itself source is a widely used unsupervised corpus for pre-training, does not contain any entity annotations related to ESE tasks. Furthermore, we use this part of data in reverse to train $\overleftarrow{\text{GPT-2}}$. Therefore, the existence of the Reverse-Wikipedia does not affect our claim that our \modelName{} is a corpus-independent ESE method.

\subsection{Compared Methods} 
We compare our method with the following ESE methods.
\begin{enumerate}
    \item Egoset~\cite{rong2016egoset}: This method uses context features and word embeddings to generate a sparse word ego network centered on seed entity. The network is used to capture words with similar semantics.
    \item SetExpan~\cite{shen2017setexpan}: This method calculates the distribution similarity to select context features from the corpus iteratively. Then SetExpan expands the entity set based on its proposed rank ensemble mechanism.
    \item SetExpander~\cite{mamou-etal-2018-term}: This method utilizes a specific classifier to predict whether the candidate entity belongs to the seed set. The input of the classifier is the context features.
    \item CaSE~\cite{yu2019corpus}: This method combines Skip-Gram context feature and embedding feature to sort all candidate entities in the corpus and then select entities to expand the entity set.
    \item Set-CoExpan~\cite{huang2020guiding}: This method automatically generates auxiliary sets which contain negative entities to explicitly alleviate the semantic drift problem.
    \item CGExpan~\cite{zhang-etal-2020-empower}: This method uses BERT to generate the class name, with the help of Hearst patterns~\cite{hearst-1992-automatic}. With the constraint of positive class name, the new entity and seed entity are in the same category, avoiding semantic drift problem. 
    \item SynSetExpan~\cite{shen-etal-2020-synsetexpan}: This method proposes to better leverages the limited supervision signals in seeds by utilizing the synonym information.
    \item ProbExpan~\cite{li2022contrastive}: Current state-of-the-art method on Wiki/APR/SE2 datasets. It designs an entity-level masked language model and conducts contrastive learning on the corpus with entity annotations.
\end{enumerate}

\begin{table}[t]
\centering
\caption{The performance (MAP@50) comparison between CGExpan and \modelName{} under different classes. }
\scalebox{1.00}{
\begin{tabular}{ccc}
\toprule
\textbf{Semantic Class} & \textbf{CGExpan} & \textbf{\modelName{}} \\
\midrule
\texttt{US States} & 0.950 & 0.947 $\downarrow$ \\
\texttt{China Provinces} & 0.662 & 0.868 $\uparrow$  \\
\texttt{Countries} & 0.965 & 1.000 $\uparrow$  \\
\texttt{Diseases} & 1.000 & 1.000 $-$ \\
\texttt{Sports Leagues} & 0.942 & 1.000 $\uparrow$  \\
\texttt{TV Channels} & 0.925 & 1.000 $\uparrow$  \\
\texttt{Parties} & 0.937 & 1.000 $\uparrow$  \\
\texttt{Companies} & 0.867 & 0.976 $\uparrow$  \\
\bottomrule
\end{tabular}
}

\label{tab:fine_result}
\end{table}

\subsection{Evaluation Metric}
The task objective of ESE is to expand a ranked list of entities that belong to the same semantic class as seed entities. Therefore, previous works widely use the Mean Average Precision at different top K positions (i.e., MAP@K) as the evaluation metric. 
Specifically, MAP@K is calculated as:
\begin{equation}
    \mbox{MAP@K}=
    \frac{1}{|Q|}\sum_{q \in Q}
    \mbox{AP}_K(R_q,G_q),
\end{equation}
where $Q$ is all the seed entity sets. And for each set $q$, the $\mbox{AP}_K(R_q,G_q)$ denotes the average precision at position $K$ with the ranked list $R_q$ and ground-truth list $G_q$. To ensure the fairness of the experiments, we set the exactly same evaluation metric used by the compared methods. Besides, the choices of $K$ values are exactly following the previous works~\cite{shen2017setexpan,zhang-etal-2020-empower,shen-etal-2020-synsetexpan}, i.e., MAP@K for K=10,20,50. All the results of our methods reported in the paper are the average results after running the model three times. 

\subsection{Implementation Details} 
All the source code of our experiments is implemented using Pytorch~\cite{paszke2019pytorch} based on the Huggingface's implementation of Transformer library\footnote{https://github.com/huggingface/transformers}~\cite{wolf-etal-2020-transformers}. 
The configuration of the two GPT-2 models in our method is the same as the default settings of Huggingface, which has 12 transformer layers with 12 attention heads and its hidden state size is 768~\cite{radford2019language}. 
The architecture of the BERT encoder we use is the same as the $BERT_{BASE}$ model~\cite{devlin-etal-2019-bert}, whose input embedding passes through 12 stacked bidirectional Transformer blocks with 768 hidden dimensions and 12 self-attention heads.
For the main results we report, we initialize the BERT encoder with the weights of BERT-wwm model~\cite{cui2019pre}. But for a fair comparison with CGExpan, we also use BERT-base-uncased to conduct experiments.
When updating the initial seed/candidate sets (Equation~\ref{update_seed} and Equation~\ref{update_cand}), we choose $\text{thr}_u = 0.65$ and $\text{thr}_l = 0.25$ as we report the results of our proposed method. When automatically generating the context patterns, we set $m$ to 130 in our experiments. As for the selection strategy of these hyper-parameters, practically, we derive a validation set for each dataset and conduct a simple hyper-parameter grid search on it. More specifically, the validation set for a dataset is generated by randomly selecting seed entities from the ground truth of each semantic class. Note that the seed entities in the validation set have no overlap with seed entities in the test set. By evaluating the model performance on the validation set under different values of hyper-parameters, we obtain the optimal hyper-parameter settings. This is standard practice for determining hyper-parameters and can be easily applied to other datasets as well as other application scenarios. 
Additionally, we will later provide a parameter study on $\text{thr}_u,\text{thr}_l$ in the experiments to analyze how these two hyper-parameters affect the performance of our proposed model.

\begin{figure}[]
\centering
\includegraphics[width=0.45\textwidth]{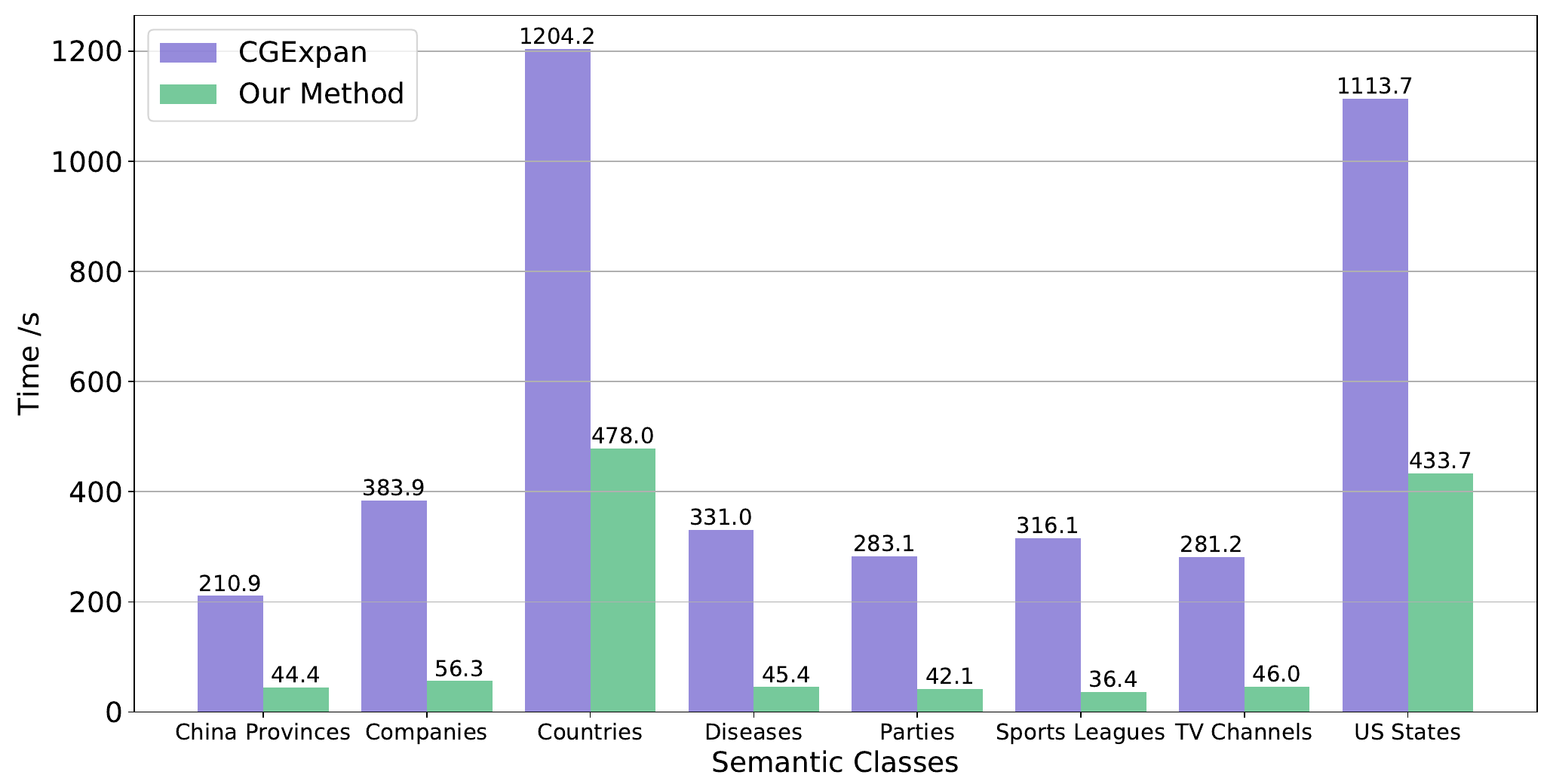}
\caption{Running efficiency analysis. For every class, we report the running time consumed by the two models when expanding 50 entities.}
\label{Efficiency_Figure}
\end{figure}

\begin{table*}[h]
    
    \centering
    \caption{CGExpan-SSE means that we perform supervision signal enhancement (SSE) module on CGExpan. \modelName{}-NoSSE means the ablation of \modelName{} without SSE module. “$\uparrow$” indicates that the corresponding method receives a performance improvement after combing with SSE module.}
    \scalebox{1.00}{
    \begin{tabular}{lccccccccc}
    \toprule \multirow{2}{*} { \textbf{Methods} } & \multicolumn{3}{c} { \textbf{Wiki} } & \multicolumn{3}{c} { \textbf{APR} }  \\ \cmidrule(r){2-4} \cmidrule(r){5-7}
    & MAP@10 & MAP@20 & MAP@50 & MAP@ 10 & MAP@20 & MAP@50  \\
    \midrule 
    CGExpan & 0.995 & 0.978 & 0.902 & 0.992 & 0.990 & 0.955 \\
    CGExpan-SSE & 0.996$\uparrow$ & 0.980$\uparrow$  & 0.916$\uparrow$  & 0.993$\uparrow$  & 0.992$\uparrow$  & 0.960$\uparrow$   \\
    \modelName{}-NoSSE & 0.997 & 0.996 & 0.956 & 0.995 & 0.992 & 0.970  \\
    \modelName{} & 1.000$\uparrow$  & 1.000$\uparrow$  & 0.974$\uparrow$  & 1.000$\uparrow$  & 1.000$\uparrow$  & 0.990$\uparrow$   \\
    \midrule
    \end{tabular}
    }
    
    \label{tab:supervision_result}
\end{table*}

\subsection{ Overall Performance} 
From Table~\ref{tab:allresult} and Table~\ref{tab:fine_result}, we can observe that:
\begin{enumerate}
    \item Our \modelName{} performs better than all baselines on all datasets and evaluation metrics. Specifically, without the participation of human-annotated corpora, \modelName{} outperforms the previous state-of-the-art corpus-based ESE models. In addition, while CGExpan uses Hearst patterns to generate class names and SynSetExpan introduces synonym information to enhance ESE, \modelName{} has better performance than them without any external information.
    \item For different values of K in MAP@K, we can see that \modelName{} performs particularly well under MAP@10 and MAP@20, where it even achieves 100\% performance on Wiki and APR. This phenomenon reflects the high quality of the ranked entity list obtained by our method, that is, the ranked list obtained by our method indeed ranks those entities with the correct semantic class preferentially.
    \item For fine-grained semantic classes, the comparison results in Table~\ref{tab:fine_result} show that \modelName{} outperforms CGExpan under most classes significantly, which reflects that our method is more generalized than CGExpan for different semantic classes. Especially for \texttt{China Provinces}/\texttt{Companies} classes, the performance of the CGExpan is significantly lower than other classes, while our method narrows the performance gap among different classes. We think that the main reason for the large difference in the performance of CGExpan for different semantic classes is that the given corpus it uses has different annotation quality for different classes of entities, which also proves the shortcoming of the corpus-based ESE methods (which is described in Section~\ref{Sec:Introduction}) and reflects the advantage of our proposed corpus-independent ESE.
\end{enumerate}

\subsection{Efficiency Analysis}
Because our proposed \modelName{} utilizes autoregressive language models to automatically generate context patterns compared to methods that use manually pre-defined patterns, it is necessary to conduct an in-depth comparative analysis of the efficiency of our method and previous work. For a fair comparison, we run CGExpan and \modelName{} under the same hardware settings. Our used CPU is the Intel(R) Xeon(R) Gold 5218R CPU @ 2.10GHz, and the GPU is the GeForce RTX 3090 with 24GB memory. We take the time that the model successfully expands 50 entities (right or wrong) as the running time for the final report of the model.

From Figure~\ref{Efficiency_Figure}, we can know that \modelName{} takes significantly less time than CGExpan to obtain the same number of expansion entities, which demonstrates that our method does not suffer from a decrease in efficiency due to our used text generation models such as GPT-2. After comparative analysis at the method design level, we think that there are two reasons for the efficiency difference shown in Figure~\ref{Efficiency_Figure}: (1) The first and mainly, in each expansion iteration, CGExpan utilizes BERT to perform the probing operations to automatically generate corresponding semantic class names, which is a very time-consuming process. But \modelName{} has pre-generated all context patterns with two GPT-2 models before entering the iterative expansion process, and used the BERT encoder to encode these context patterns, so there is no need to run these large-scale models such as GPT-2 and BERT iteratively. (2) The second, kinds of ranking strategies designed in CGExpan are more complex than ours. CGExpan needs to rank both the generated class names and the selected entities. The complex ranking operation increases the time consumed by CGExpan to a certain extent.

\subsection{Ablation Studies}

\subsubsection{Effectiveness of Supervision Signal Enhancement Module} To verify the effectiveness of our designed supervision signal enhancement module, we analyze the performance impact of the supervision signal enhancement module on \modelName{} and CGExpan. 
From Table~\ref{tab:supervision_result} we can see that after combining with the supervision signal enhancement module, all evaluation metrics of CGExpan and \modelName{}-NoSSE on the two datasets have been improved. These steady performance improvements prove that our proposed supervision signal enhancement module is helpful to alleviate the sparse supervision problem in ESE. Furthermore, the performance advantage of \modelName{}-NoSSE over CGExpan shows that even without the participation of the supervision signal enhancement module, our model still has a competitive performance.

\subsubsection{Parameter Sensitivity Analysis} 
\begin{figure}[h]
\centering
\setlength{\abovecaptionskip}{0.5em}
\subfigure[Wiki-$\text{thr}_{l}$] 
{ \label{wikiu} 
\includegraphics[height = 0.39 \columnwidth, width=0.46\columnwidth]{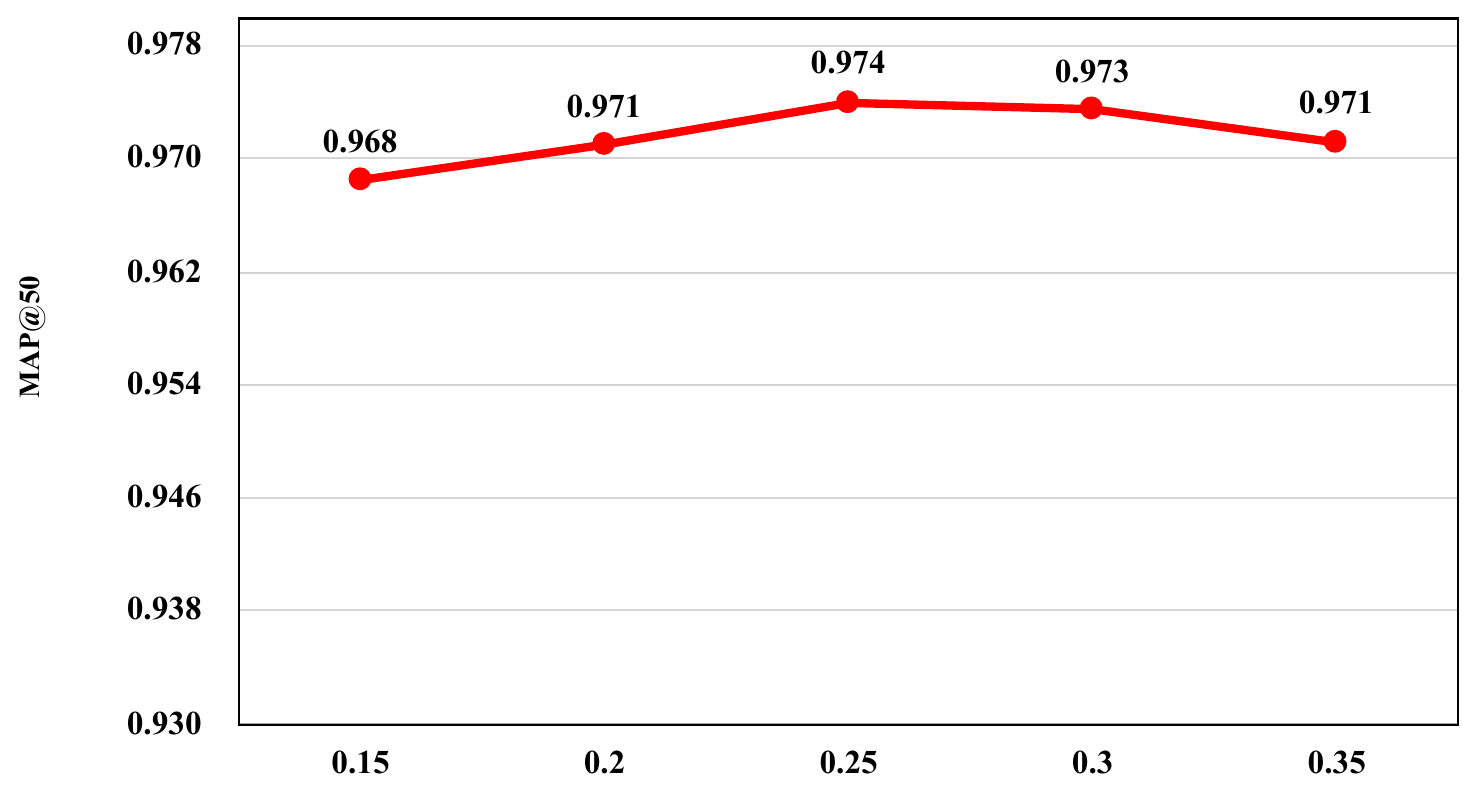} 
} 
\subfigure[Wiki-$\text{thr}_{u}$] 
{ \label{wikil} 
\includegraphics[height = 0.39 \columnwidth,width=0.46\columnwidth]{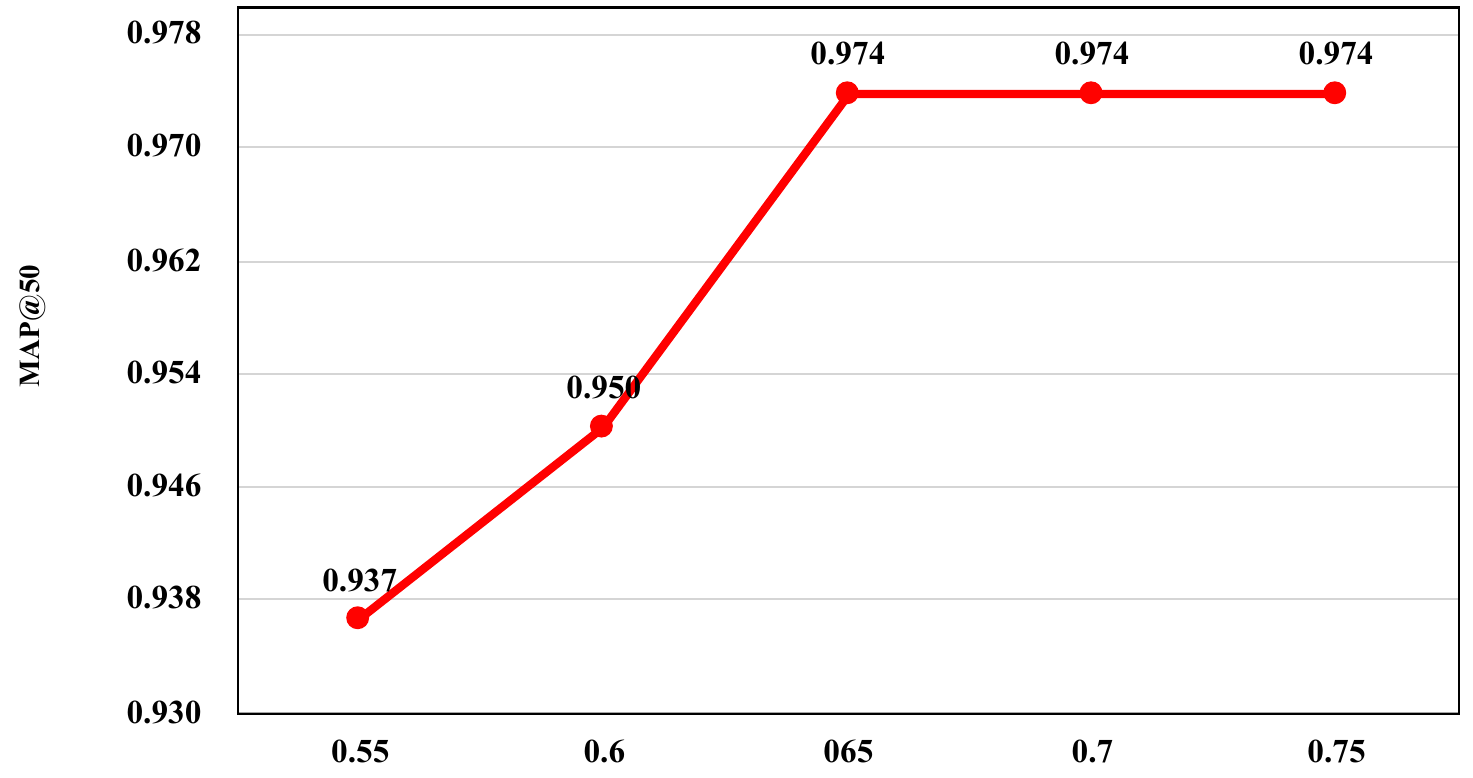} 
} 

\subfigure[APR-$\text{thr}_{l}$] 
{ \label{apru} 
\includegraphics[height = 0.39 \columnwidth, width=0.46\columnwidth]{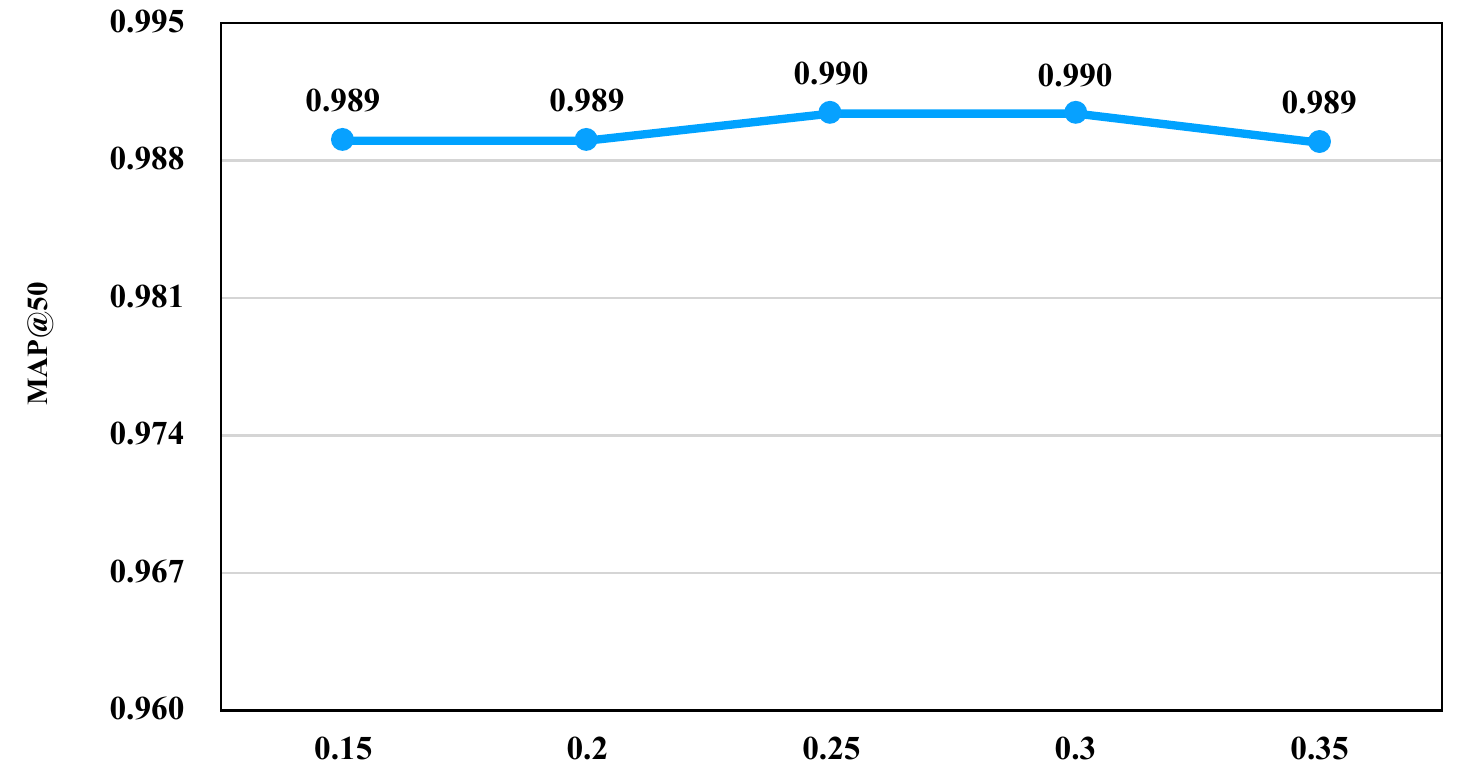} 
} 
\subfigure[APR-$\text{thr}_{u}$] 
{ \label{aprl} 
\includegraphics[height = 0.39 \columnwidth, width=0.46\columnwidth]{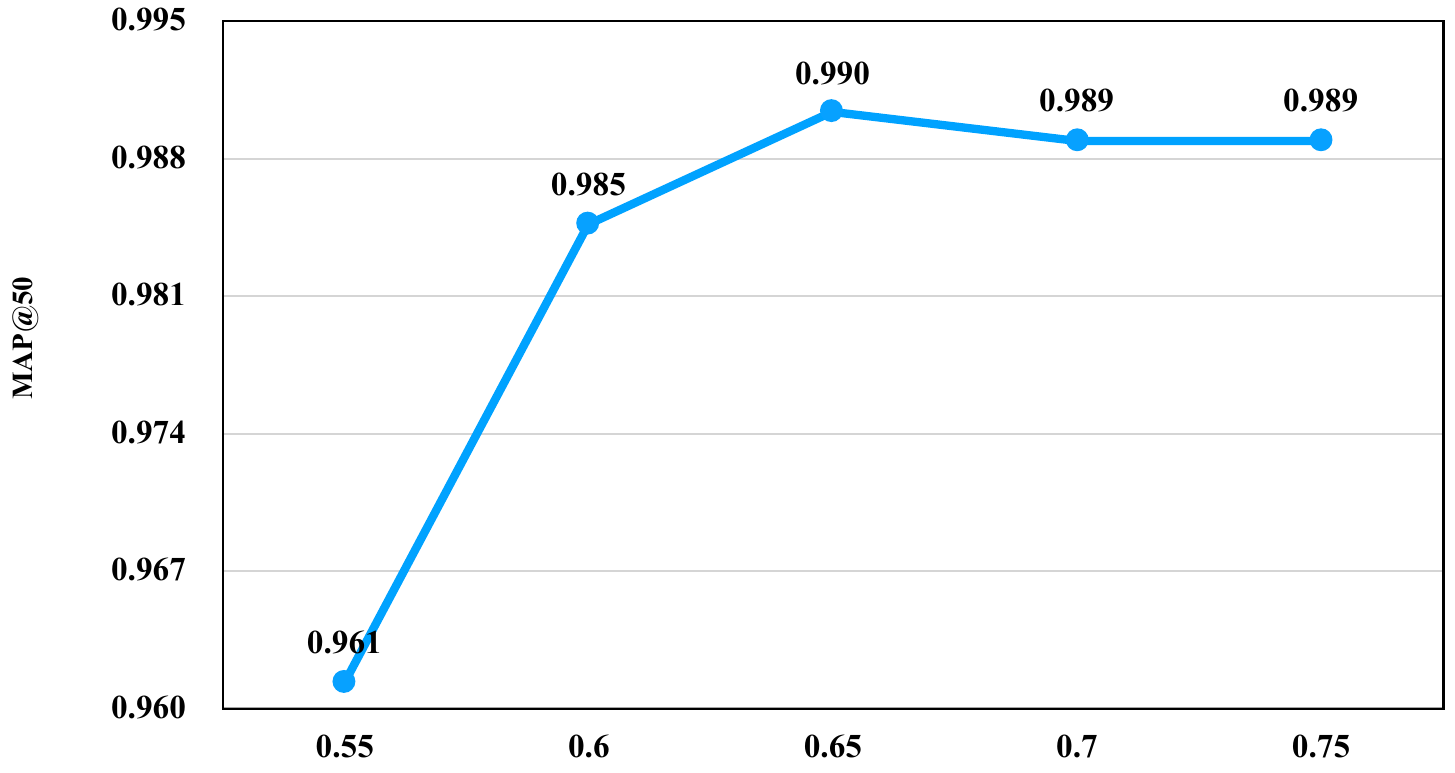} 
}
\caption{Parameter sensitivity analysis of $\text{thr}_{u}$, $\text{thr}_{l}$ in \modelName{}. The MAP@50 of the state-of-the-art models on Wiki and APR are 0.926 and 0.960.} 
\label{tab:sensitive} 
\end{figure}

In our proposed \modelName{} framework, there are two key hyper-parameters (i.e., $\text{thr}_{u}$ and $\text{thr}_{l}$) for updating the initial seed/candidate sets to enhance the supervision signal for our model. Therefore, we further analyze how these two hyper-parameters affect the overall performance of our model. Specifically, we report the model performance with various values of $\text{thr}_{u}$, $\text{thr}_{l}$ on different datasets. Note that because our method performs very stable on MAP@10 and MAP@20, the performance remains at 1.000 for almost every hyper-parameter value, so we only use MAP@50 results for demonstration. As shown in Figure~\ref{tab:sensitive}, the performance of \modelName{} in all cases is higher than the previous state-of-the-art performance, which reflects the stable competitive performance of our proposed model.

Figure~\ref{wikiu} and Figure~\ref{apru} show that as the $\text{thr}_{l}$ changes, the model performance does not change very significantly, usually just to three or four decimal places. In \modelName{}, the direct reason why we want to use the $\text{thr}_{l}$ is that we can use $\text{thr}_{l}$ as a threshold to reduce the candidate set size, thereby reducing the search space in the expansion process. The second advantage of using $\text{thr}_{l}$ is that the number of expansion iterations can be decreased, thereby reducing the computational consumption and running time. Therefore, when the value of $\text{thr}_{l}$ changes in a relatively small interval (e.g., $[0.15, 0.35]$), it will only cause the candidate set size to change, because the entities involved in the values of these smaller $\text{thr}_{l}$ must be entities that are far apart in semantic space from the seed entities, the presence or absence of these entities has little effect on the model performance.

From Figure~\ref{wikil} and Figure~\ref{aprl}, we can see that as the $\text{thr}_{u}$ increases, the performance of the model first increases and then tends to remain unchanged. As described in Section~\ref{Sec:Seed_Entities_Enhancement}, we use $\text{thr}_{u}$ as a threshold to select entities from the initial candidate set to expand the initial seed set. Therefore, when the value of $\text{thr}_{u}$ is too small, some entities with low similarity to the seed entities will be wrongly introduced into the seed set, thereby introducing noise for the subsequent expansion process, which naturally reduces the expansion performance. Conversely, when the value of $\text{thr}_{u}$ increases to a certain value, the overall expansion performance of the model no longer changes significantly, which indicates that when we set $\text{thr}_{u}$ as a value within a suitable interval, the model performance will be not sensitive for the value of $\text{thr}_{u}$. This phenomenon is consistent with the design motivation that we originally want to select entities with high similarity from the candidate set to expand the initial seed set, thereby enhancing the supervision signal.

\begin{figure}[h]
\centering
\setlength{\abovecaptionskip}{0.5em}
\subfigure[Wiki-$\text{m}$] 
{ \label{wikim} 
\includegraphics[height = 0.39 \columnwidth, width=0.46\columnwidth]{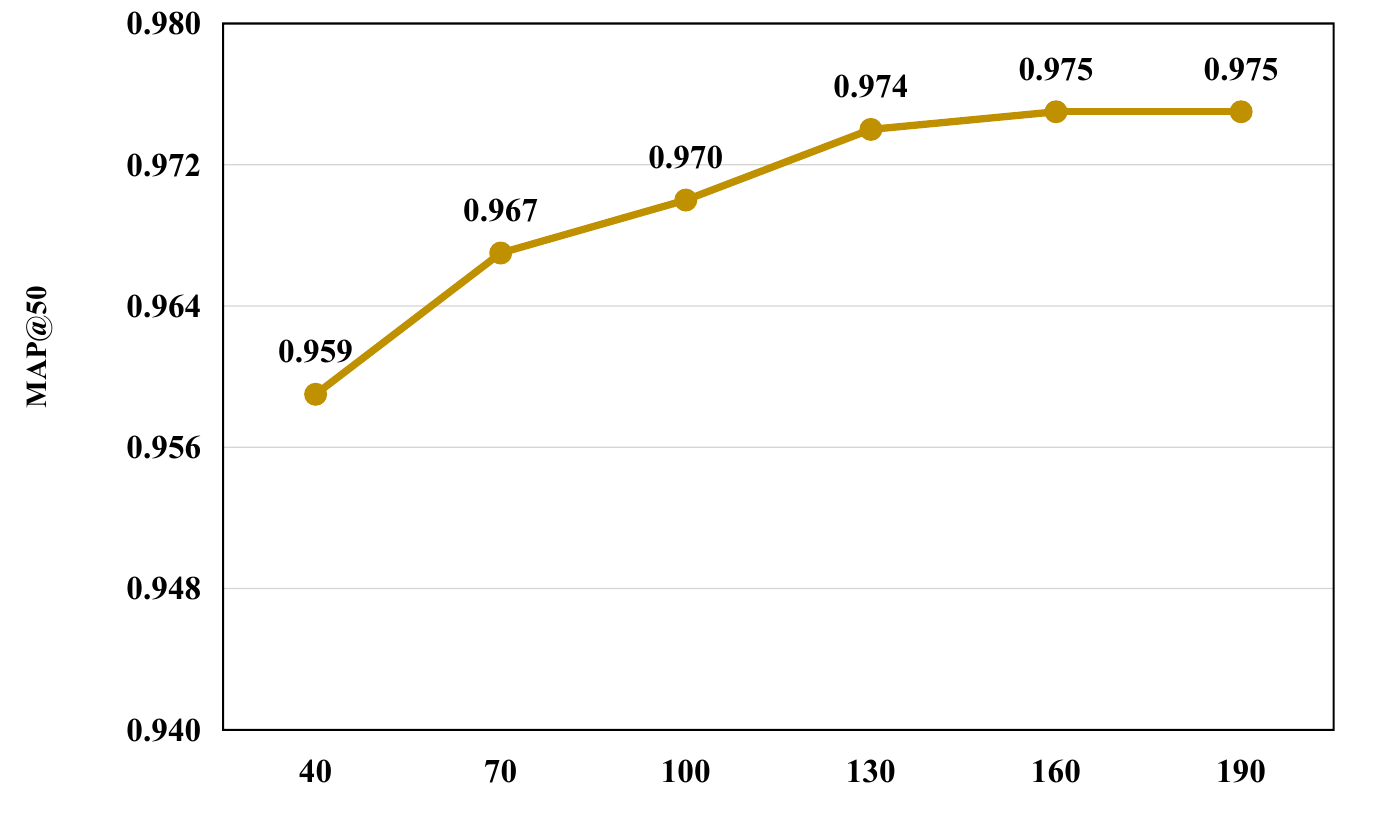} 
} 
\subfigure[APR-$\text{m}$] 
{ \label{aprm} 
\includegraphics[height = 0.39 \columnwidth,width=0.46\columnwidth]{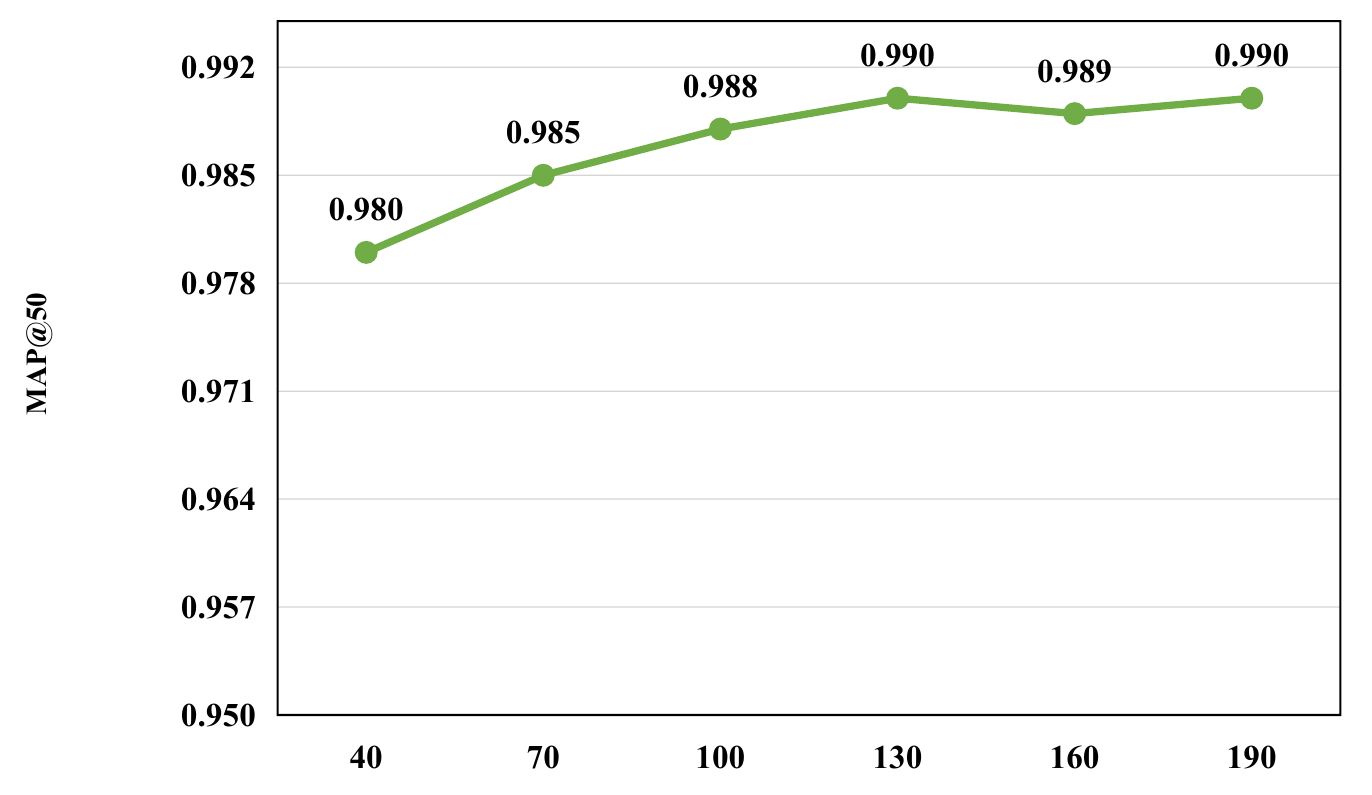} 
} 
\caption{Parameter sensitivity analysis of $\text{m}$ in \modelName{}. The MAP@50 of the state-of-the-art models on Wiki and APR are 0.926 and 0.960.}
\label{tab:m_sensitive} 
\end{figure}

Additionally, we also analyze how the context patterns number (i.e., $m$) affects the performance. Figure~\ref{wikim} and Figure~\ref{aprm} show that as the $m$ increases, there is a slight boost in performance. Benefiting from the strategy of increasing supervised signals, the performance has already reached competitive results with $m$ of 40. Note that under various values of $m$, the performance of \modelName{} is better than the previous state-of-the-art performance, which proves that our method is not very sensitive to the value of $m$. To achieve a balance between efficiency and effectiveness, we recommend a setting of 130 for $m$ in practice.

\begin{table}[h]
\centering
\caption{Rouge score between generated context patterns and various corpora.}
\label{tab:pattern_rouge}
\begin{tabular}{lccc}
\toprule
\textbf{Dataset} & \textbf{Rouge-1} & \textbf{Rouge-2} & \textbf{Rouge-L} \\ \midrule
Wiki & 0.368  &0.203  &0.361  \\
SE2 & 0.413  &0.279  &0.482  \\
Reverse-Wikipedia & 0.431  &0.302  &0.534   \\
\bottomrule
\end{tabular}
\end{table}

\begin{table}[h]
\small
\centering
\caption{Perplexity and Distinct n-gram (Dist-n) of generated contexts and reference contexts.}
\label{tab:context}
\begin{tabular}{lcccc}
\toprule
\textbf{} & \textbf{Perplexity$\downarrow$} & \textbf{Dist-1$\uparrow$} & \textbf{Dist-2$\uparrow$} & \textbf{Dist-3$\uparrow$} \\ \midrule
Generated & 38.921  & 0.247  & 0.547 & 0.705 \\
Wikipedia & 36.263  & 0.171  & 0.485 & 0.653  \\
\bottomrule
\end{tabular}
\end{table}

\begin{table*}[h]
\centering
\caption{Context patterns generated for corresponding entities. The 1st and 3rd columns are the prev-text and next-text automatically generated by $\protect\overleftarrow{\text{GPT-2}}$ and $\protect\overrightarrow{\text{GPT-2}}$, respectively.}
\scalebox{0.90}{
\begin{tabular}{|c|c|c|}
\hline
\textbf{Generated Prev-Text}  & \textbf{Entity} & \textbf{Generated Next-Text}   \\ \hline
\makecell[c]{... were used to act counter-clockwise \\ at Nanchong , or at ,} & \emph{“Sichuan”}  & \makecell[c]{, China . The first two of these\\ are the most common ...} \\ \hline
... a football competition in & \emph{“England”} & \makecell[c]{, which was held from \\ 1892 to 1896 ...}\\ \hline
\makecell[c]{... a useful clinical sign in distinguishing \\ malaria from other causes of} & \emph{“fever”} & \makecell[c]{, but it is not a reliable \\ diagnostic test for malaria ...} \\ \hline
\makecell[c]{... He was purchased by the Rajasthan Royals \\ for the 2008 and 2009 seasons of} & \emph{“Indian Premier League”} & \makecell[c]{(IPL) and was a part of the \\ team that won the title ...}   \\ \hline
\makecell[c]{... Alfonso Foods was a fast food hamburger\\ drive-in restaurant chain based in} & \emph{“California”} & \makecell[c]{. It was founded in 1946 \\ by Alfredo ...} \\ \hline
\makecell[c]{... after creating her Facebook page,\\ Bean also joined}  & \emph{“Twitter”}  &\makecell[c]{ and Instagram . She is \\ a regular contributor ...} \\ \hline 
\makecell[c]{... the company design websites for\\ many clients including,} & \emph{“Microsoft”} & \makecell[c]{, Apple, Google, Facebook\\, Amazon, eBay ...} \\ \hline
\makecell[c]{... a huge area stretching from \\ Yunnan , a southwest into }& \emph{“China”} & \makecell[c]{'s far west , to the \\ Tibetan plateau ...}  \\ \hline
\makecell[c]{... HeTe was selected as a member\\ and joined the} & \emph{“Reform party”} & \makecell[c]{in 1983 . He was a \\ founding member ...} \\ \hline
\makecell[c]{... who is best known for starring \\ in a drama with} & \emph{“TVB”} & \makecell[c]{ and the Hongkong \\ Film Production ...} \\ \hline
\end{tabular}
}
    
    \label{tab:case_patterns_result}
\end{table*}

\subsubsection{Context patterns Analysis}
To verify the quality of context patterns, we analyze the Rouge score between generated context patterns and various related corpus. 
For each initial seed entity set, we get the sentences that contain the seed entity from the corpus and regard these sentences as the reference. At the same time, generated context patterns of the seed entity are obtained via our context pattern generation module. 
To measure the overlap between generated context patterns and reference sentences, we calculate the average Rouge-1, Rouge-2, and Rouge-L scores between them.
Specifically, for each generated context containing an entity, Rouge scores are computed between the context and reference contexts containing the same entity, and the highest Rouge scores among various reference contexts represent the similarity scores between the generated context and the corpus. Similarity scores for all generated contexts are averaged as the entire overlap between generated contexts and reference contexts.


From the results of Wiki and SE2 in Table~\ref{tab:pattern_rouge}, the low Rouge scores indicate that our generated context patterns do not have a high degree of coincidence with the corpus used by traditional corpus-based ESE methods, so our method does not have an information leakage problem.
And the low Rouge scores of Reverse-Wikipedia reflect that \modelName{} generates diverse contexts related to entities, rather than simply copying what it has seen in the pre-training corpus.


Furthermore, we compare the quality of the generated contexts to those of Wikipedia, evaluating them based on fluency and diversity. 
We employ perplexity~\cite{liu2016not} to measure the fluency and Distinct n-gram (Dist-n)~\cite{li2016diversity} to measure the diversity. GPT-Neo-1.3B~\cite{black2022gpt} is utilized for computing perplexity.
For fair comparison in Dist-n, we sample the same number of contexts in Wikipedia as generated contexts.
The results presented in Table~\ref{tab:context} demonstrate that \modelName{}'s generated contexts are not only similar in fluency to those in Wikipedia, but also exhibit higher diversity. When it comes to long-tail entities, Wikipedia only contains a limited number of contexts available, making it challenging to expand uncommon entities for the ESE task. Meanwhile, \modelName{} is capable of generating a comparable number of contexts for each entity with high diversity, making it easier to expand uncommon but semantically relevant entities.


\begin{table*}[h]
\small
\centering
\caption{According to Table~\ref{tab:fine_result}, we select the \texttt{China Provinces} and \texttt{Companies} classes because they lead to poor CGExpan's performance, and we present the results of the \texttt{US States} class to analysis in which case will \modelName{} have poor performance. We mark the wrong entities in red.}
\scalebox{0.90}{
\begin{tabular}{|c|c|c|c|c|}
\hline
\textbf{Seed Entity Set}                                                                                              & \multicolumn{2}{c|}{\textbf{CGExpan}} & \multicolumn{2}{c|}{\textbf{\modelName{}}}  \\ \hline
\multirow{10}{*}{\begin{tabular}[c]{@{}c@{}}\texttt{China Provinces} \\ \\ \{\emph{“Anhui”},\emph{“Fujian”},\emph{“Hunan”}\}\end{tabular}} 
& 1           & \emph{“Hubei”}          & 1            & \emph{“Henan”}           \\ \cline{2-5} 
& 2           & \emph{“Guangdong”}          & 2            & \emph{“Shandong”}             \\ \cline{2-5} 
& 3           & \emph{“Jiangxi”}         & 3            & \emph{“Guangdong province”}              \\ \cline{2-5} 
&          & ...        &           & ...          \\ \cline{2-5} 
& 29       & \emph{“\textcolor{red}{Nanjing}”}         & 29           & \emph{“Yunnan”}        \\ \cline{2-5} 
& 30          & \emph{“\textcolor{red}{Wuhan}”}         & 30           & \emph{“Ningxia”}     \\ \cline{2-5} 
& 31          & \emph{“Inner Mongolia”}        & 31           & \emph{“Jiangxi”} \\ \cline{2-5} 
& 32          & \emph{“\textcolor{red}{Hangzhou}”}         & 32           & \emph{“Henan province”} \\ \cline{2-5} 
& 33          & \emph{“\textcolor{red}{Guangzhou}”}         & 33           & \emph{“Liaoning province”} \\ \cline{2-5} 
&          & ...        &           & ...            \\ \hline
\multirow{10}{*}{\begin{tabular}[c]{@{}c@{}}\texttt{Companies} \\ \\ \{“Myspace”,“Twitter”,“Youtube”\}\end{tabular}} 
& 1           & \emph{“Flickr”}          & 1            & \emph{“facebook”}                 \\ \cline{2-5} 
& 2           & \emph{“\textcolor{red}{wikipedia}”}          & 2            & \emph{“Livejournal”}           \\ \cline{2-5} 
& 3           & \emph{“AOL”}          & 3            & \emph{“Discogs”}         \\ \cline{2-5} 
&          & ...        &           & ...          \\ \cline{2-5} 
& 29          & \emph{“\textcolor{red}{Safari}”}         & 29           & \emph{“AIM”}  \\ \cline{2-5} 
& 30          & \emph{“CNet”}         & 30   & \emph{“Lego”}    \\ \cline{2-5} 
& 31          & \emph{“\textcolor{red}{Second Life}”}         & 31           & \emph{“\textcolor{red}{Mtv}”}        \\ \cline{2-5} 
& 32          & \emph{“\textcolor{red}{Microsoft Outlook}”}         & 32           & \emph{“Macromedia”}         \\ \cline{2-5} 
& 33          & \emph{“\textcolor{red}{Mozilla Thunderbird}”}         & 33           & \emph{“Microsoft”}       \\ \cline{2-5} 
&          & ...        &           & ...          \\ \hline
\multirow{10}{*}{\begin{tabular}[c]{@{}c@{}}\texttt{US States} \\ \\ \{“Ohio”,“Mississippi”,“Indiana”\}\end{tabular}} 
& 1           & \emph{“Illinois”}          & 1            & \emph{“Missouri”}                 \\ \cline{2-5} 
& 2           & \emph{“{Arizona}”}          & 2            & \emph{“Arkansas”}           \\ \cline{2-5} 
& 3           & \emph{“California”}          & 3            & \emph{“Iowa”}         \\ \cline{2-5} 
&          & ...        &           & ...          \\ \cline{2-5} 
& 46          & \emph{“{Alaska}”}         & 46           & \emph{“Nevada”}  \\ \cline{2-5} 
& 47          & \emph{“Vermont”}         & 47   & \emph{“\textcolor{red}{Kansas State University}”}   \\ \cline{2-5} 
& 48          & \emph{“{Kansas State}”}         & 48           & \emph{“\textcolor{red}{Orange County}”}        \\ \cline{2-5} 
& 49          & \emph{“{Hawaii}”}         & 49           & \emph{“\textcolor{red}{Mississippi River}”}         \\ \cline{2-5} 
& 50          & \emph{“\textcolor{red}{Mexico}”}         & 50           & \emph{“Louisvile”}       \\  \hline
\end{tabular}
}
    
    \label{tab:case_entities_result}
\end{table*}

\subsection{Case Studies}

\subsubsection{Context Pattern Generation}
Table~\ref{tab:case_patterns_result} shows some cases of our context pattern generation module for several entities from different semantic classes. We can see that the texts generated by the two GPT-2 models in opposite directions are of high quality, both in terms of relevance and fluency. Especially for the prev-text generated by $\overleftarrow{\text{GPT-2}}$, please kindly note that in practice $\overleftarrow{\text{GPT-2}}$ generates the prev-text from right to left under the guidance of the entity, which fully proves that our pre-trained reverse GPT-2 model is able to generate reverse text. To the best of our knowledge, our work firstly demonstrates that it is completely feasible to train GPT-2 with reversed corpus so that it has the ability of reverse generation.

Moreover, as can be seen from the cases in Table~\ref{tab:case_patterns_result}, the context patterns generated by our model inadvertently introduce many related entities, which strengthens the representation of semantic classes. We believe this is one of the reasons why our automatically generated context patterns can greatly improve the performance of the model.

\subsubsection{Generated Patterns Guided Expansion} 

From the results of class \texttt{China Provinces} and \texttt{Companies} shown in Table~\ref{tab:case_entities_result}, it can be seen that our model effectively avoids some entities with very similar semantic class when expanding. 
It is because CGExpan uses pre-defined patterns to generate class names, and these pre-defined patterns cannot distinguish fine-grained semantic classes (e.g., \emph{“Anhui”} and \emph{“Nanjing”} both belong to the \texttt{Location} class, but \emph{“Nanjing”} belongs to \texttt{China City} and  \emph{“Anhui”} belongs to \texttt{China Province} on a more granular level). 
The comparison of these two semantic classes further illustrates the advantage of our proposed corpus-independent ESE paradigm, that is our context pattern generation module can perceive fine-grained semantic classes and automatically generate high-quality patterns accordingly.

In addition, it is also necessary to analyze in which case \modelName{} does not perform well, so as to find the direction of further improvement in the future. From Table~\ref{tab:fine_result}, we can see that our model does not perform as well as CGExpan when faced with the \texttt{US States} class, so we further compare and analyze this semantic class in our case study. 
Interestingly, we find that \modelName{} sometimes incorrectly expands some nested entities (e.g., \emph{“Kansas State University”} and \emph{“Mississippi River”}). We think this phenomenon may be related to the way we automatically generate context patterns. If we use \emph{“Mississippi”} in the seed set as the guide text to generate context patterns, then $\overrightarrow{\text{GPT-2}}$ is likely to generate the word “river” as the next-text, this will lead to \modelName{} wrongly judging that \emph{“Mississippi River”} is an entity with high similarity with the seed entity set. Therefore, it is a worthwhile research direction to study how to avoid introducing noise in the context pattern generation process. We suggest that it is possible to control the context patterns generated by generative language models by optimizing decoding strategies that more closely match ESE scenarios.

\section{Conclusion and Future Work}
\label{Sec:Conclusion}
In this paper, we design a simple yet effective strategy for updating the initial seed/candidate sets and apply it to the supervision signal enhancement module to alleviate the long-standing problem of sparse supervision in ESE.
More importantly, we propose to promote the ESE task by automatically generating context patterns for entities with the help of two separate GPT-2 models in opposite directions. This pioneering attempt frees our model from the dependence on manually annotated corpus and brings the ESE task into the new corpus-independent paradigm.
Empirical results on three widely used benchmarks show the competitive performance of our method.
In the future, we will study how to automatically measure the quality of the generated context patterns to further enhance our method. 

\appendices


\ifCLASSOPTIONcompsoc
  \section*{Acknowledgments}
\else
  \section*{Acknowledgment}
\fi

Yinghui Li, Shulin Huang, and Xinwei Zhang realize the development methodology and the creation of models. Qingyu Zhou, Yangning Li, and Ruiyang Liu are responsible for the investigation process, experimental verification, and the writing and revision of the paper. Yunbo Cao and Hai-Tao Zheng are responsible for the management of the team, the planning of the scientific research progress, and the financial support. Ying Shen is responsible for checking the correctness and innovation of the proposed method, as well as the writing and revision of the manuscript.


\ifCLASSOPTIONcaptionsoff
  \newpage
\fi



\bibliographystyle{IEEEtran}
\bibliography{IEEEabrv}
%



%
\begin{IEEEbiography}[{\includegraphics[width=1in,height=1.25in,clip,keepaspectratio]{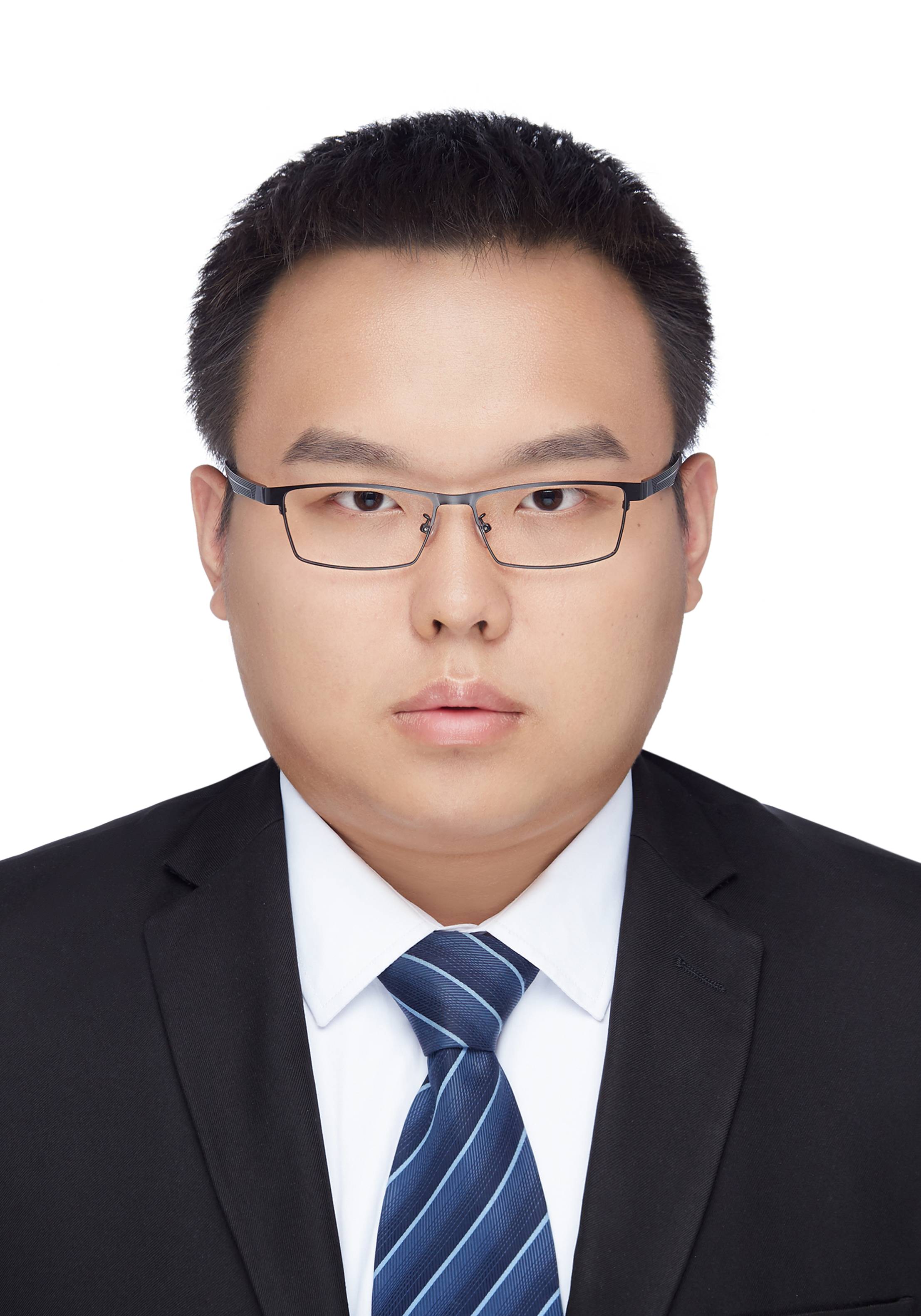}}]{Yinghui Li} received the BEng degree from the Department of Computer Science and Technology, Tsinghua University, in 2020. He is currently working toward the PhD degree with the Tsinghua Shenzhen International Graduate School, Tsinghua University. His research interests include natural language processing and deep learning.
\end{IEEEbiography}

\begin{IEEEbiography}[{\includegraphics[width=1in,height=1.25in,clip,keepaspectratio]{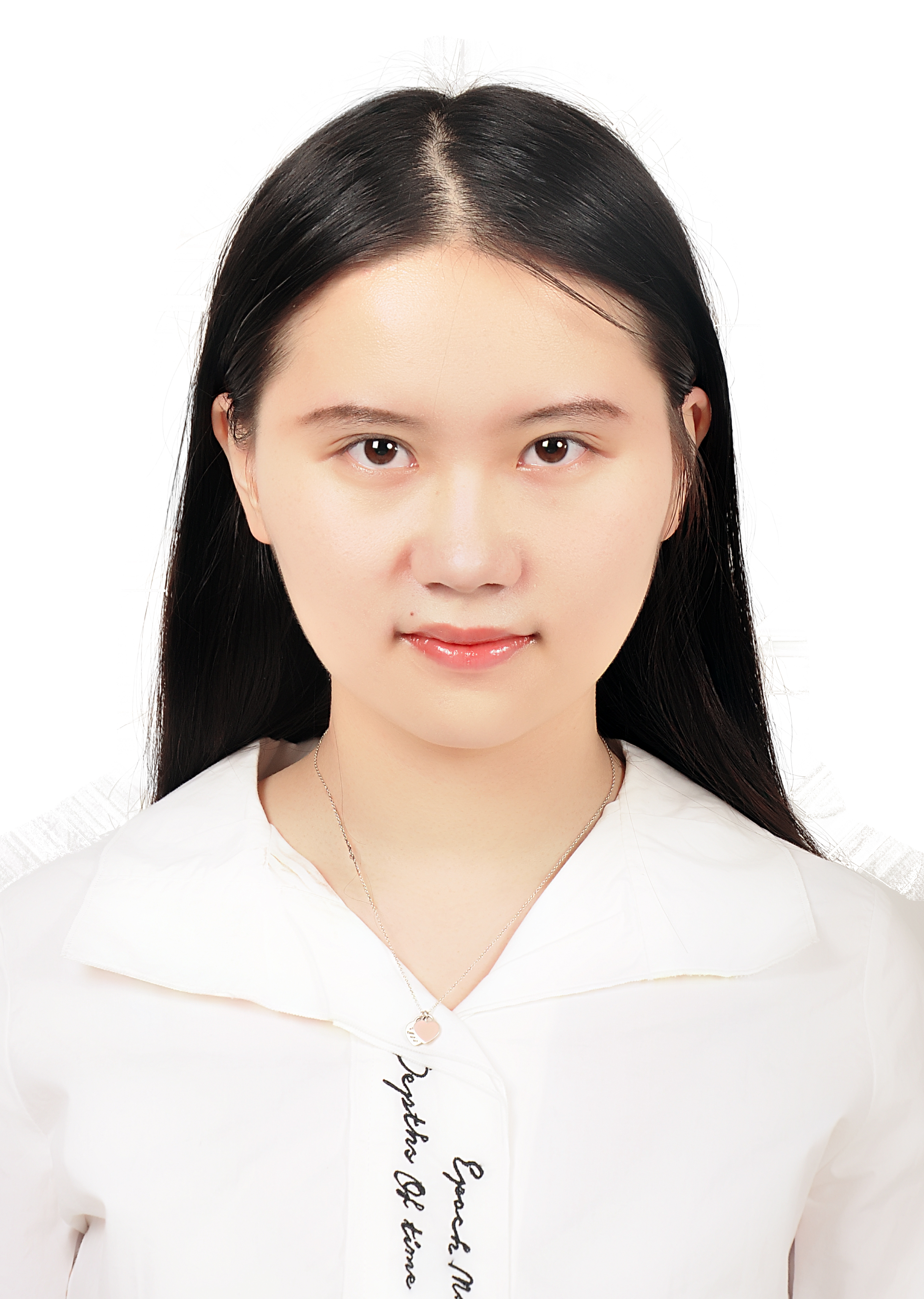}}]{Shulin Huang} received the BEng degree from the Department of Computer Science and Engineering, Northeastern University, in 2021. She is currently working toward Master’s degree with the Tsinghua Shenzhen International Graduate School, Tsinghua University and mainly focusing on natural language processing.
\end{IEEEbiography}

\begin{IEEEbiography}[{\includegraphics[width=1in,height=1.25in,clip,keepaspectratio]{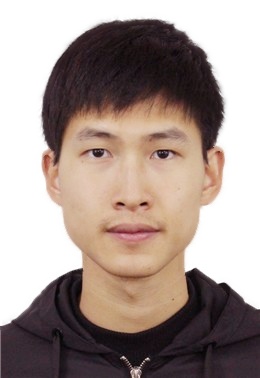}}]{Xinwei Zhang} received the BEng degree from the Department of Intelligence Science and Technology, Sun Yat-sen University, in 2021. His research interests include natural language processing and deep learning.
\end{IEEEbiography}

\begin{IEEEbiography}[{\includegraphics[width=1in,height=1.25in,clip,keepaspectratio]{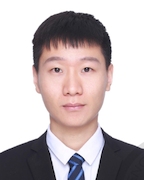}}]{Qingyu Zhou} received his PhD degree in a joint program of Harbin Institute of Technology and Microsoft Research Asia. He is currently a Senior Researcher at Tencent, Beijing, China. His research interests include text summarization, natural language generation, dialogue system and educational applications of NLP.
\end{IEEEbiography}

\begin{IEEEbiography}[{\includegraphics[width=1in,height=1.25in,clip,keepaspectratio]{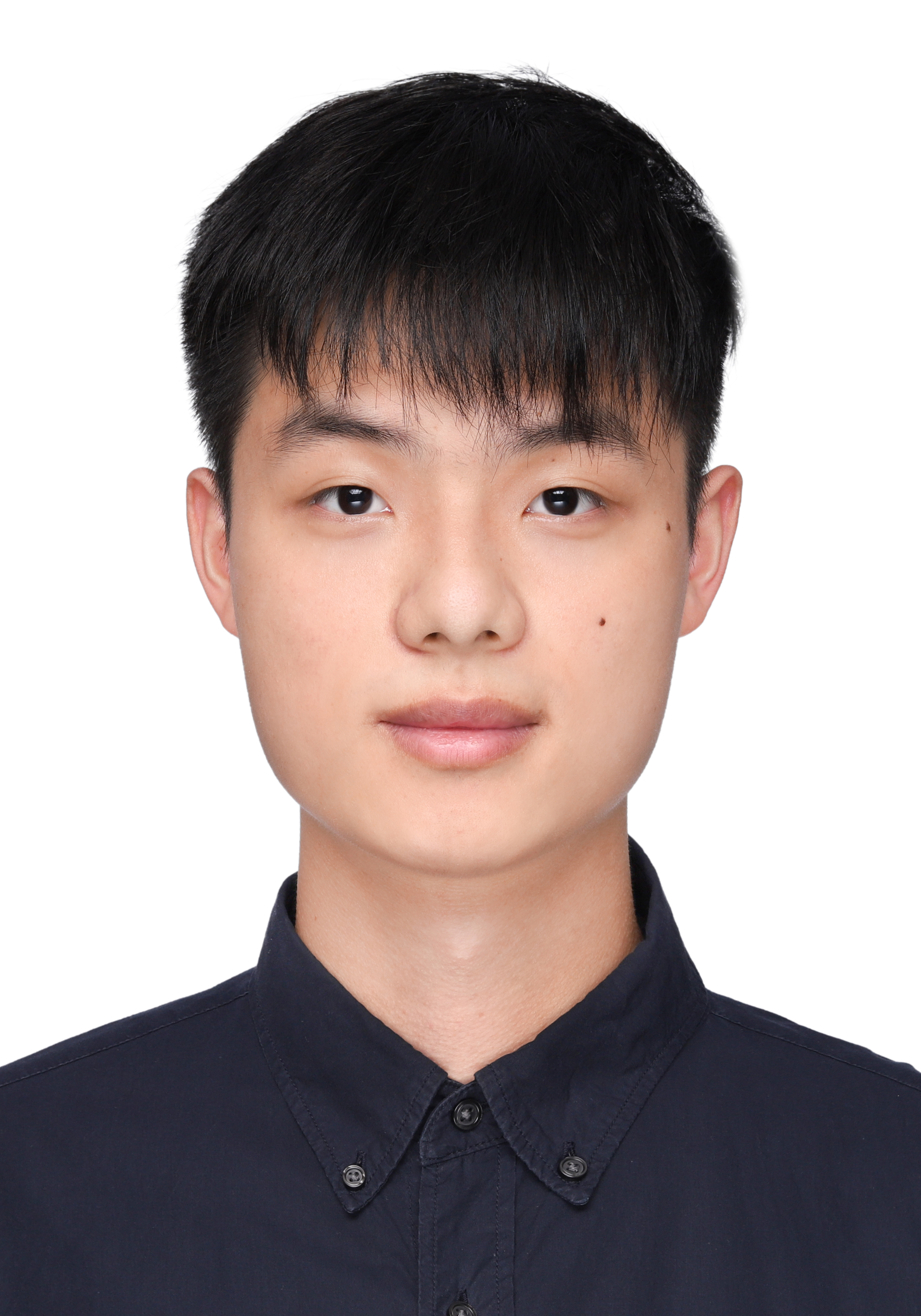}}]{Yangning Li} received the BEng degree from the Department of Computer Science and Technology, Huazhong University of Science and Technology, in 2020. He is currently working toward a Master's degree with the Tsinghua Shenzhen International Graduate School, Tsinghua University. His research interests include natural language processing and data mining.
\end{IEEEbiography}

\begin{IEEEbiography}[{\includegraphics[width=1in,height=1.25in,clip,keepaspectratio]{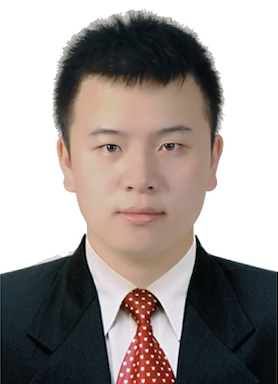}}]{Ruiyang Liu} received the BEng degree from the Department of Computer Science and Technology, Tsinghua University, in 2020. He is currently working toward Master’s degree with the Department of Computer Science and Technology, Tsinghua University and mainly focusing on deep learning, computer vision, and discrete differential equation.
\end{IEEEbiography}

\begin{IEEEbiography}[{\includegraphics[width=1in,height=1.25in,clip,keepaspectratio]{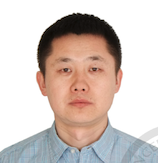}}]{Yunbo Cao} received his PhD degree in computer science from the Shanghai Jiao Tong University, in 2011. He is currently the head of the NLP Center of Tencent Cloud Xiaowei and mainly focuses on various applications of NLP.
\end{IEEEbiography}

\begin{IEEEbiography}[{\includegraphics[width=1in,height=1.25in,clip,keepaspectratio]{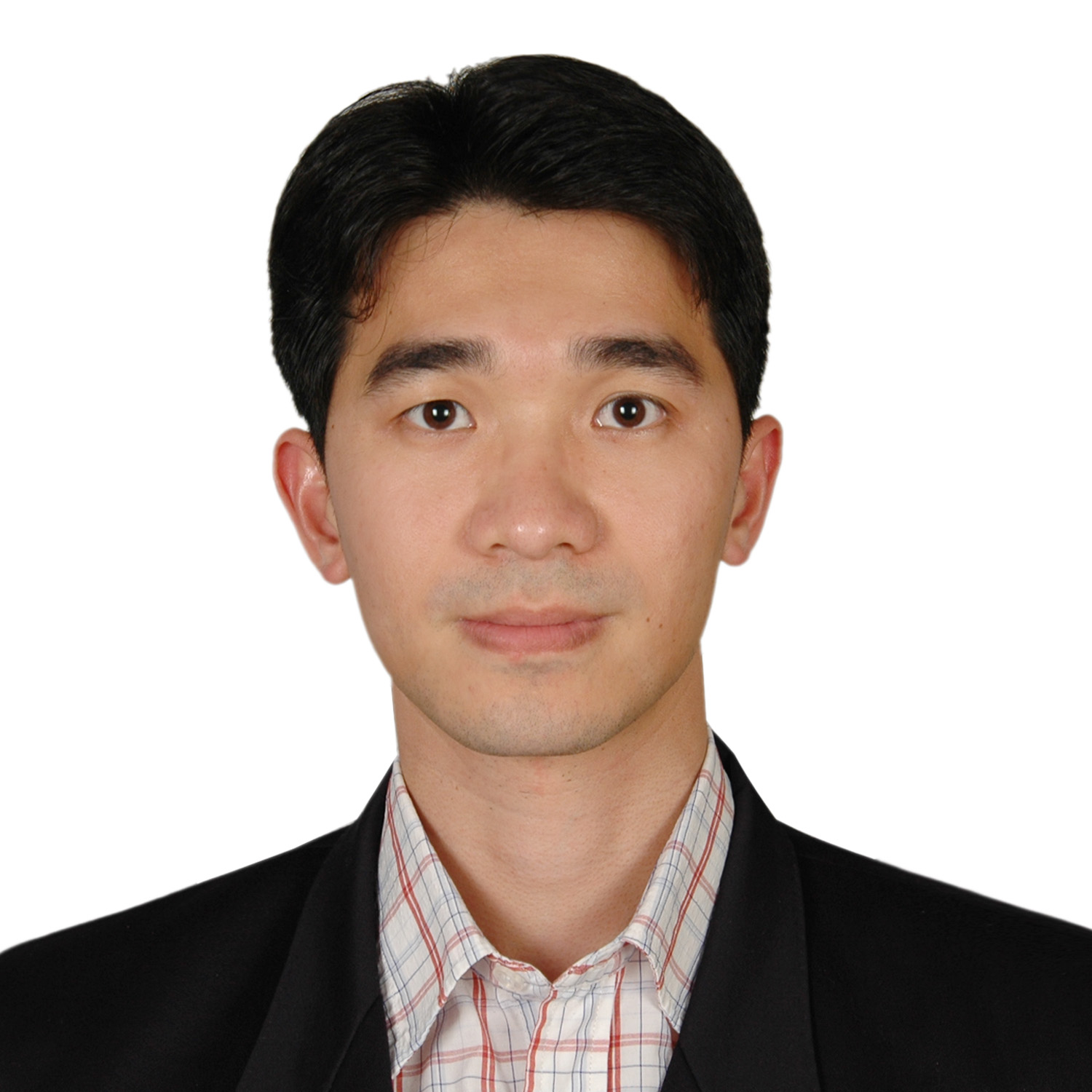}}]{Hai-Tao Zheng} received the bachelor’s and master’s degrees in computer science from the Sun Yat-Sen University, China, and the PhD degree in medical informatics from Seoul National University, South Korea. He is currently an associate professor with the Shenzhen International Graduate School, Tsinghua University, and also with Peng Cheng Laboratory. His research interests include web science, semantic web, information retrieval, and machine learning.
\end{IEEEbiography}

\begin{IEEEbiography}[{\includegraphics[width=1in,height=1.25in,clip,keepaspectratio]{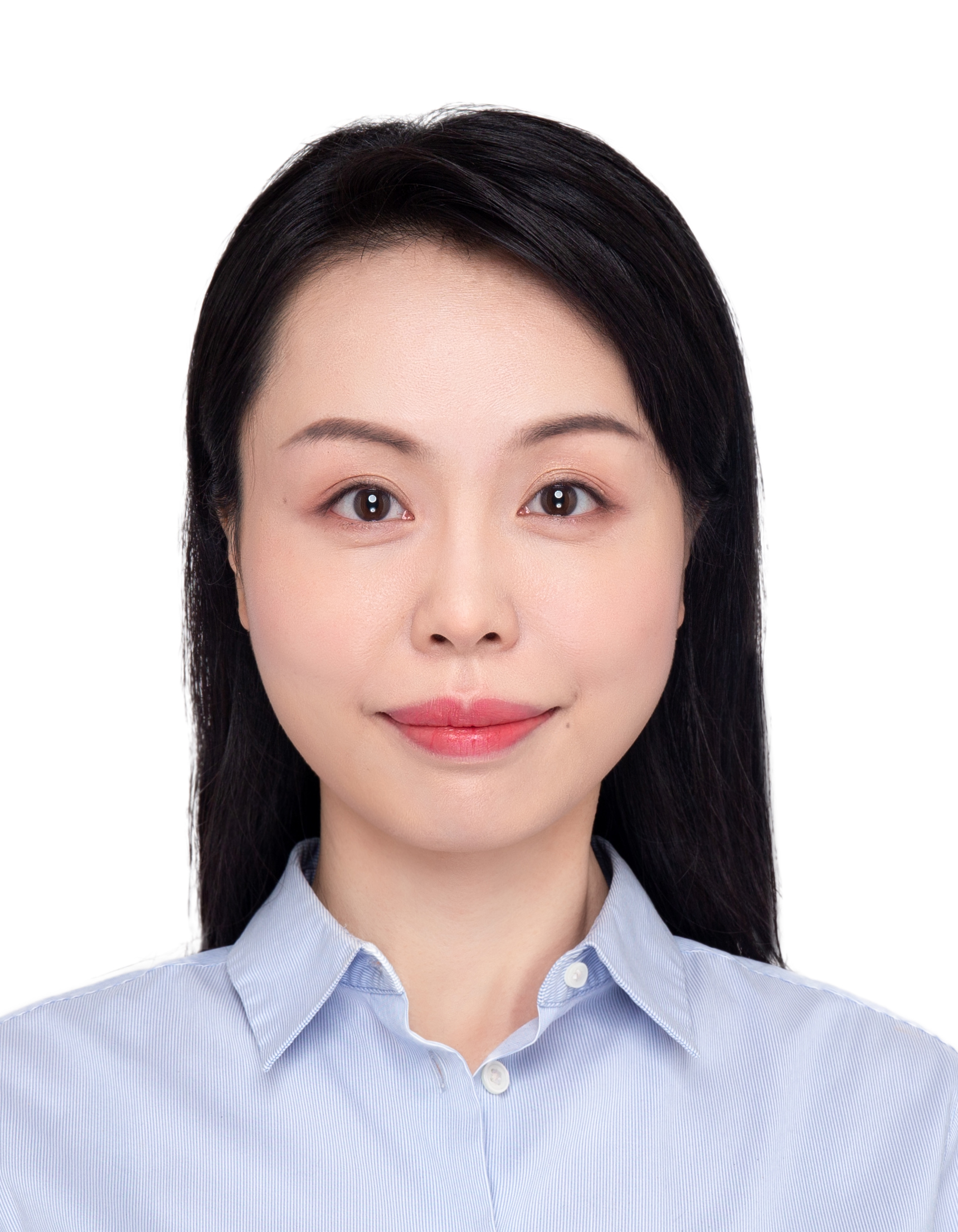}}]{Ying Shen} received the PhD degree in computer science from the University of Paris Ouest Nanterre La Défense, France and the Erasmus Mundus master's degree in natural language processing from the University of Franche-Comt, France and the University of Wolverhampton, U.K. She is currently an associate professor with the School of Intelligent Systems Engineering, Sun Yat-Sen University. Her research interests include natural language processing and deep learning.
\end{IEEEbiography}







\end{document}